\documentclass[10pt,twocolumn,letterpaper]{article}

\usepackage[pagenumbers]{cvpr}              %

\definecolor{cvprblue}{rgb}{0.21,0.49,0.74}
\usepackage[breaklinks,colorlinks,allcolors=cvprblue]{hyperref}

\def\paperID{} %
\def\confName{CVPR}
\def\confYear{2026}

\usepackage[utf8]{inputenc} %
\usepackage[T1]{fontenc}    %
\usepackage{url}            %
\usepackage{booktabs}       %
\usepackage{amsfonts}       %
\usepackage{nicefrac}       %
\usepackage{microtype}      %
\usepackage[table,dvipsnames]{xcolor}         %
\usepackage{subcaption}
\usepackage{wrapfig}
\usepackage{placeins}
\usepackage{graphicx}
\usepackage{tcolorbox}
\usepackage[shortlabels]{enumitem}
\usepackage{longtable}
\usepackage{titletoc}
\usepackage[accsupp]{axessibility} 

\definecolor{linkcolor}{rgb}{0,0,0.55} %

\definecolor{voa}{RGB}{232,87,26} 
\definecolor{vob}{RGB}{233,154,9} 

\newcommand{\ourDS}{VisualOverload}
\newcommand{\vlms}{VLMs}
\newcommand{\paragraphx}[1]{\noindent\textbf{#1}}

\title{\includegraphics[height=0.40cm]{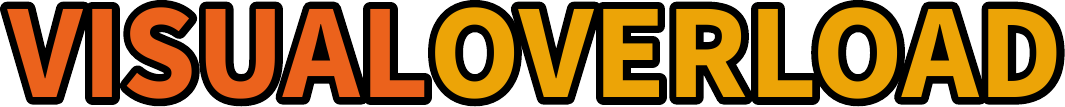}\\ Probing Visual Understanding of \vlms~in \textit{Really} Dense Scenes}

\author{%
  \textbf{Paul Gavrikov}\textsuperscript{\textcolor{voa}{\ensuremath{1,4}}}\!\quad\textbf{Wei Lin}\textsuperscript{\textcolor{vob}{\ensuremath{2}}}\!\quad\textbf{M. Jehanzeb Mirza}\textsuperscript{\textcolor{voa}{\ensuremath{3}}}\!\quad\textbf{Soumya Jahagirdar}\textsuperscript{\textcolor{vob}{\ensuremath{4}}}\\ 
\textbf{Muhammad Huzaifa}\textsuperscript{\textcolor{vob}{\ensuremath{4}}}\!\!\quad\textbf{Sivan Doveh}\textsuperscript{\textcolor{voa}{5}}\!\!\quad\textbf{Serena Yeung-Levy}\textsuperscript{\textcolor{voa}{\ensuremath{5}}}\!\!\quad\textbf{James Glass}\textsuperscript{\textcolor{voa}{\ensuremath{3}}}\!\!\quad\textbf{Hilde Kuehne}\textsuperscript{\textcolor{vob}{\ensuremath{4,\!6}}} \\\\
   \textsuperscript{\textcolor{voa}{1}}Independent Researcher\!\quad  \textsuperscript{\textcolor{vob}{2}}JKU Linz\!\quad   \textsuperscript{\textcolor{voa}{3}}MIT CSAIL\!\quad  \textsuperscript{\textcolor{vob}{4}}Tübingen AI Center\!\quad  \textsuperscript{\textcolor{voa}{5}}Stanford\\ \textsuperscript{\textcolor{vob}{6}}MIT-IBM Watson AI Lab}

\begin{document}
\maketitle
\begin{abstract}
Is basic visual understanding really solved in state-of-the-art \vlms? We present \ourDS, a slightly different visual question answering (VQA) benchmark comprising 2,720 question–answer pairs, with privately held ground-truth responses. Unlike prior VQA datasets that typically focus on near global image understanding, \ourDS~challenges models to perform simple, knowledge-free vision tasks in densely populated (or, \textit{overloaded}) scenes. Our dataset consists of high-resolution scans of public-domain paintings that are populated with multiple figures, actions, and unfolding subplots set against elaborately detailed backdrops. We manually annotated these images with questions across six task categories to probe for a thorough understanding of the scene. We hypothesize that current benchmarks overestimate the performance of \vlms, and encoding and reasoning over details is still a challenging task for them, especially if they are confronted with densely populated scenes. Indeed, we observe that even the best model (\textit{o3}) out of 37 tested models only achieves $19.6 \%$ accuracy on our hardest test split and overall $69.5 \%$ accuracy on all questions. Beyond a thorough evaluation, we complement our benchmark with an error analysis that reveals multiple failure modes, including a lack of counting skills, failure in OCR, and striking logical inconsistencies under complex tasks. Altogether, \ourDS~exposes a critical gap in current vision models and offers a crucial resource for the community to develop better models.

\noindent\textbf{Benchmark:} \url{https://paulgavrikov.github.io/visualoverload}\
\end{abstract}

\begin{figure}
    \centering    
    \includegraphics[width=\linewidth]{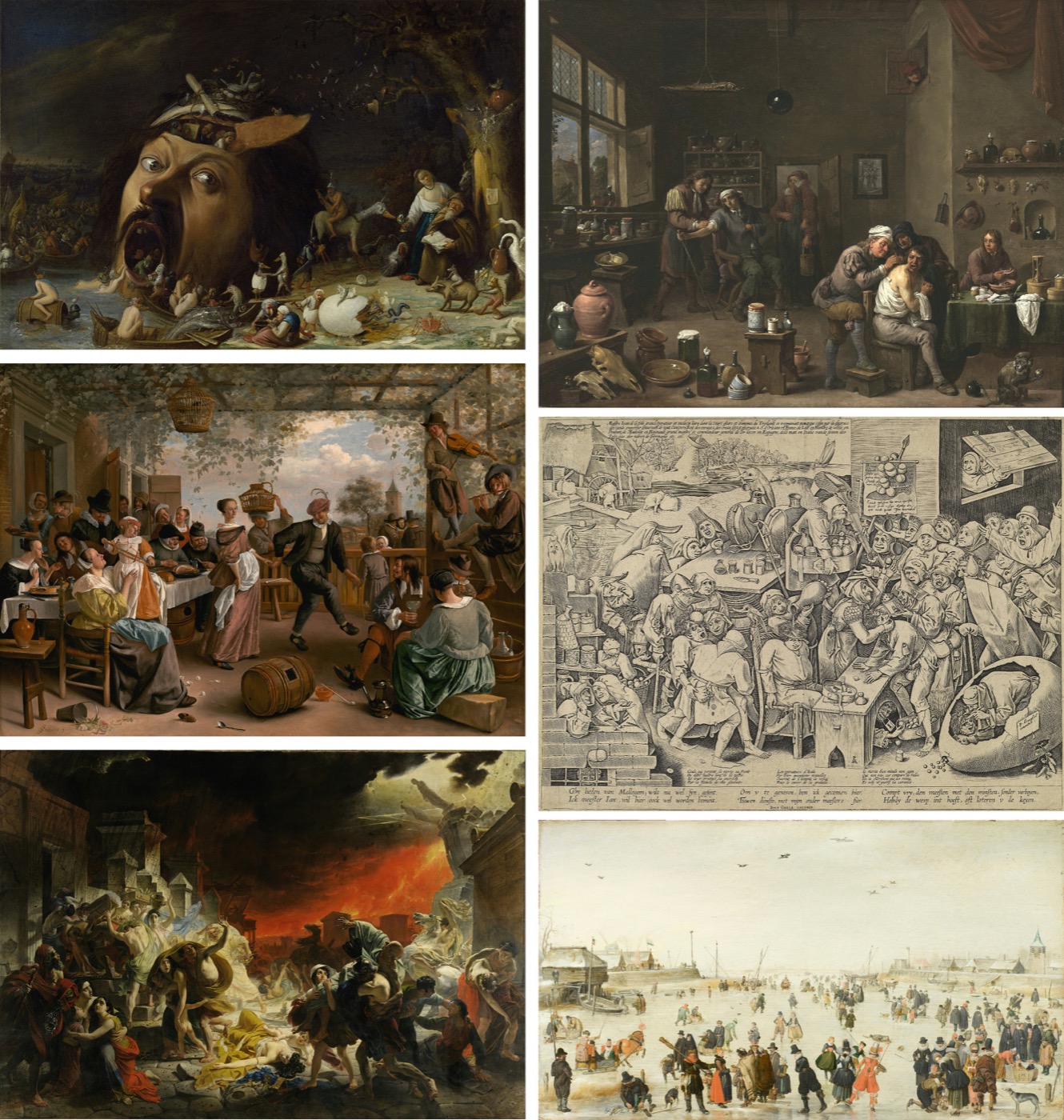}
    \caption{\textbf{Example scenes from \ourDS.} All artworks (public domain) in our benchmark show incredibly rich scenes filled with details, which we utilize to test the perception of state-of-the-art \vlms. Please zoom in for details.}
\label{fig:example_scenes}
\end{figure}
\begin{figure*}[!h]
\centering
\includegraphics[width=\linewidth]{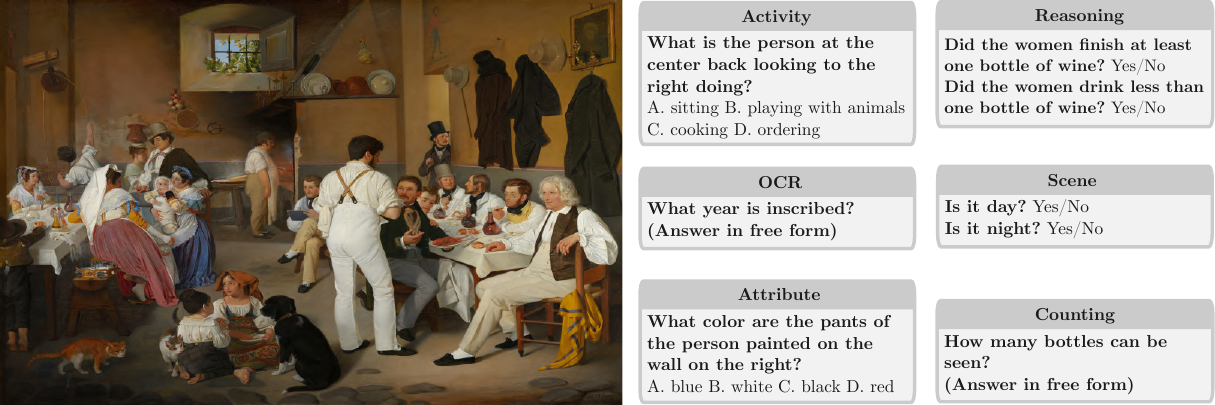}
\caption{ 
\textbf{Example questions from \ourDS.} Our benchmark consists of images displaying densely populated scenes paired with handcrafted questions (multiple-choice and free-form) covering six core vision tasks. All yes/no questions are paired with questions asking for a logical opposite question to decrease the random chance and to provide an additional signal for measuring \textit{logical consistency}.}
\label{fig:example_questions}
\end{figure*}

\section{Introduction}\label{sec:introduction}

Visual question answering (VQA) \citep{antol2015vqa,Goyal_2017_CVPR,Agrawal_2018_CVPR} has emerged as a common benchmark for image understanding in \vlms. Recent state-of-the-art models achieve surprisingly strong results on established VQA datasets \citep{seedbench2023,mmvet2024}, suggesting that basic forms of visual understanding might already be ``solved''. In turn, several benchmarks have shifted from generic image understanding towards the probing of domain-specific knowledge \citep{yue2024mmmu,phan2025humanitysexam}. %

However, \textit{have today's \vlms~really solved core vision tasks?} We argue that current benchmarks are poor indicators of this, as most of them fail to capture the complexity of real-world applications, where safety and reliability depend on fine-grained perception in dense and high-resolution scenes. %
Current benchmarks instead emphasize simple foreground reasoning in low-resolution scenes \citep{seedbench2023,omnibench2025,mmvet2024} or utilize higher resolution only for needle-in-a-haystack-like retrieval tasks \citep{vstar2023,otterhd,shi2025scaling}, falling short of testing broader capabilities, and potentially overestimating performance. 

Instead, we expect that model performance will severely drop ``under pressure'', and modulate this through the angle of visual complexity and dense--\textit{visually overloaded}--scenes. We motivate our analysis by suggesting that the vision representation and its multi-modal alignment are a bottleneck in modern \vlms. Encoders are designed to compress visual input from spatial to a semantic space, where empirical risk minimization encourages the retention of necessary features. This process imposes an inherent upper bound on fine-grained perception and attention to features that are not commonly seen during training. While random noise illustrates an extreme case of this, we expect sufficiently densely populated scenes to already trigger these limits. 

To verify our expectations, we introduce a new dataset explicitly designed to probe image understanding in dense and high-resolution scenes. Our dataset comprises 150 high-resolution scans of artworks featuring highly dense scenes, along with 2,720 manually curated question–answer pairs spanning six fundamental tasks of visual comprehension: activity recognition, attribute recognition, counting, optical character recognition (OCR), visual reasoning, and global scene classification (see~\cref{fig:example_scenes} for examples of images and~\cref{fig:example_questions} for example questions about one scene). Unlike many prior benchmarks that recycle existing image datasets, all of our images are newly sourced from public domain artworks, resulting in a fresh source of data and free of copyright concerns. 

Our empirical study of 37 \vlms~reveals that state-of-the-art models, while often competent at global scene classification, consistently struggle in fine-grained recognition in dense scenes. To better characterize these challenges, we split our benchmark into three difficulty levels (easy, medium, hard), calibrated by average model performance. Even the strongest model we tested (o3) achieves only $19.6 \%$ accuracy on the hardest split and $69.5 \%$ overall, underscoring the difficulty of the benchmark and the underlying challenge.

Finally, we conduct a detailed error analysis and uncover striking failures: for instance, we observe strong failures in counting tasks for high ground-truth values and in OCR tasks requiring precise textual recognition, such as the recognition of typos. Furthermore, we observe that models frequently provide logically inconsistent answers to logically opposite paired questions, with this instability intensifying as the complexity of such queries increases. Such inconsistencies sometimes even degrade performance to random or even sub-random baselines, suggesting that these models rely heavily on shortcuts rather than robust reasoning. Taken together, these findings highlight the urgent need for benchmarks like ours that reflect the realities of dense, high-resolution perception and reveal fundamental limitations of current \vlms.

We summarize our contributions as follows:
\begin{itemize}%
\item We introduce a new benchmark for VQA in dense, high-resolution (\textit{visually overloaded}) scenes. Our benchmark contains 2,720 manually curated question–answer pairs across six fundamental categories (activity recognition, attribute recognition, counting, OCR, visual reasoning, and global scene classification) as described in \cref{sec:creating}. Ground truths are held private to avoid target leakage. All images are sourced entirely from public domain artwork collections to provide a fresh image dataset free of copyright issues.
\item We evaluate a range of state-of-the-art models in \cref{sec:experiments} and show that, while they perform well on global scene classification, they struggle significantly in fine-grained understanding in dense settings, particularly for counting and OCR. We provide a three-level difficulty split, calibrated by average model performance, showing that even the strongest tested model (o3) reaches only $19.6 \%$ accuracy on the hardest split.
\item We perform a detailed task-level error analysis in \cref{sec:error_analysis}, uncovering systematic inconsistencies and shortcut biases that further hinder robust performance in visually overloaded settings.
\end{itemize}

\section{The \ourDS~Benchmark}\label{sec:creating}

Our goal is to create a benchmark that tests basic image recognition skills that we expect to be present in any frontier models. However, unlike many previous benchmarks, we design our benchmark around fine-grained recognition in dense scenes to stress test the vision encoders' representation and their multi-modal alignment. In the following subsections, we discuss the dataset curation (\cref{subsec:bench:curation}), evaluation process (\cref{subsec:bench:eval}), and discuss differences to other benchmarks in detail (\cref{subsec:bench:comparison}).

\subsection{Dataset Curation}\label{subsec:bench:curation}

\paragraphx{Image collection.}
We collected 150 high-resolution digitizations of paintings, curated from collections held by museums around the world and made available through Google Arts \& Culture (see~\cref{fig:example_scenes} for examples). We specifically selected paintings that depict visually complex scenes--densely composed narratives filled with numerous figures, actions, and subplots, often unfolding simultaneously within richly detailed environments. As complexity is hard to quantify \cite{forsythe2011predicting,marin2014examining,mahon2023minimumdescriptionlengthclustering,lin2025what}\footnote{In~\cref{subsec:appendix:complexity} we show an experiment to estimate the complexity of all samples in our dataset via VLMs.}, we picked artworks that tend to overwhelm our eyes and demanded significant time and attention to fully absorb their intricate details. We only selected paintings in the public domain, automatically granted for artworks where the original creators passed away more than 100 years ago \cite{RichardStim2023Jan}.

Due to the inherent complexity of the scenes, the images in the dataset are typically of extreme resolution and exceed 4K resolution ($3840 \times 2160$ pixels). To standardize the dataset, we downsampled all images to match the nearest number of pixels in 4K, while preserving their original aspect ratios. 28 images were originally below 4K resolution and were therefore not downsampled; however, all images remain above Full HD resolution ($1920 \times 1080$ pixels).

\paragraphx{Question annotation.}
Six human annotators manually annotated the resized images with questions and answer options. The annotators were instructed to generate questions that are clearly formulated and specific, leaving no ambiguity about the information being requested. To avoid language priors, the questions are also explicitly mandated to be grounded in the content of the accompanying image and should not be answered from text alone \citep{Zhang_2016_CVPR,Goyal_2017_CVPR,agrawal2016analyzing,Agrawal_2018_CVPR,cadene2019rubi}. In addition, we restricted questions to probe for details that can be directly observed or reasonably inferred from the image, excluding any question–answer pairs based on beliefs or subjective interpretations. Finally, we requested questions to be solvable without external or expert knowledge beyond a basic level of everyday ``world’’ knowledge, as we are only concerned with image understanding in this work (unlike prior art-based benchmarks \citep{Garcia2020aqua,Becattini2023VISCOUNTH}).

We employ two answer formats: multiple-choice and freeform. Multiple-choice questions either offer four options, with only one correct answer, or are binary yes/no questions. We pair each of the latter kind of questions with a logical opposite (\eg, \textit{"Is it day?"} and \textit{"Is it night?"}) \citep{Zhang_2016_CVPR}. This not only helps calibrate against random guessing but also provides an additional signal for identifying logical inconsistencies in generated responses (see \cref{sec:error_analysis}). For selected tasks, we use freeform answers to raise the level of difficulty (see below).

Our annotated questions each fall into one of the following six categories, resulting in approximately 18 questions per image:
\begin{itemize}%
    \item \textbf{Activity recognition ($N{=}150$):} multiple-choice questions about actions or activities occurring in the scene. These questions will refer to a single or a group of subjects, typically paired with a constraint. For instance, \textit{"What is the person dressed in brown at the front of the table in the leftmost house doing?"}.
    \item \textbf{Attribute recognition ($N{=}149$):} multiple-choice queries about the color attribute of objects are typically paired with a constraint probing for spatial, attribute, or activity recognition. For instance, \textit{"What is the color of the left-most ship flag?"}.
    \item \textbf{Counting ($N{=}559$):} freeform inquiries about details that involve determining the number of objects present. The questions may be related to the entire scene or spatially constrained, requiring mild visual reasoning to provide a correct answer. For instance, \textit{"How many roses are lying on the floor?"}.
    \item \textbf{OCR ($N{=}118$):} freeform queries about written text in the image. Languages include English, Latin, Chinese, Dutch, and Greek. Some questions are specified to probe for parts of the text, which can be seen as a mild form of text reasoning, \eg, \textit{"What is the last name of the signature?"}, or require some minimal visual reasoning efforts, \eg, \textit{"What does the word below the main character read?"}.
    \item \textbf{Reasoning ($N{=}356$):} multiple-choice queries that require a medium to high load of visual reasoning to be answerable. In principle, we expect that a "chain of thought" is necessary to provide a correct answer. For instance, these questions may require functional or intent understanding, distance or path estimation, light- or wind-source estimation, occupancy detection, and numerical comparisons based on the image's content. Some example questions are: \textit{"Do you have to cross the water to reach the two windmills on the right?"}, \textit{"I am allergic to seafood, is all of the food on the table safe for me?"}, or \textit{"Does capital punishment appear to be legal in this scene?"}.
    \item \textbf{Scene classification ($N{=}1388$):} multiple-choice queries about the overall scene or setting of the image. These questions typically do not require a fine-grained understanding or complex visual reasoning of the scene. Thus, we primarily see these samples as a test for shallow image understanding, and we expect all models to perform well on them. Yet, we still observe that some models struggle with them. For instance, \textit{"Are there animals in the scene?"}.
\end{itemize}

\paragraphx{Quality control.}
After annotation, we evaluated 37 \vlms~on our dataset and manually verified the correctness of ground truths if the question was only solved by a small number of models. 
Furthermore, we evaluated the performance of 3 of the strongest models from our leaderboard (InternVL3-38B, Qwen2.5-VL 32B, LLaVA-OV 72B) on our dataset while ablating the image (\ie, a blind question answering) to probe for hidden biases due to linguistic cues in the question or answer options of multiple-choice questions. We detected a number of questions where all 3 models were able to answer the question without seeing the image. We then prompted Gemini 2.5 Pro to detect potential language biases in each question (see~\cref{appendix:language_bias} for the prompt) and removed instances with severe biases, such as cases where the correct answer was an oddity or was implied by the context of the question.
Please note that this is not necessary for freeform answers (counting, OCR) or binary questions, which are self-balanced by their logical opposites. The final ``blind'' performance on the remaining questions is shown in Appendix \cref{tab:language_bias}. Overall, our quality control resulted in a reduction of blind performance to near-chance baselines for most tasks. However, we still observe above-random performance for the attribute recognition and counting tasks. These gains stem primarily from statistical irregularities in the distribution of ground-truth answers (\eg, small object counts being more frequent). Such distributional priors are unavoidable in real-world datasets and do not confer a generalizable shortcut that undermines evaluation. In practice, models must still extract and process visual content to achieve strong performance on all of our tasks.

\paragraphx{Difficulty splits.} We divide our questions into three difficulty levels--easy, medium, and hard--based on the ratio of correct responses by all models described in \cref{sec:experiments}. The thresholds are defined by the percentage of correct responses per question: $[0,20]$ for hard, $(20,90)$ for medium, and $[90,100]$ for easy.

\subsection{Evaluation Process}\label{subsec:bench:eval}

\paragraphx{Metrics.}
We rely on the average accuracy as the principal metric for our benchmark, scored over all questions, each difficulty split, as well as each task category. We define an answer as accurate if it precisely matches the ground truth label. For binary questions, we measure pair-wise accuracy, and score a pair as correct if both questions are correct, and false otherwise.

\paragraphx{Answer extraction.}
Although our default prompts (see~\cref{subsec:appendix:prompts}) aim to constrain output format, \vlms~do not always follow these instructions. To address this, we apply simple heuristic-based preprocessing to extract and normalize responses across tasks.

For multiple-choice questions, we detect the option letter and map it to the corresponding label, or directly match the label when possible.
For counting questions, we extract either numeral or lexical integer forms, defaulting to the last-mentioned integer if multiple candidates appear.
For OCR tasks, we extract the relevant text, then normalize it by removing diacritics, punctuation, and spacing, converting to lowercase, and replacing `V' with `U' and `J' with `I' to reduce ambiguity in Latin texts.

\paragraphx{Evaluation server.}
To prevent test leakage into future \vlms, we hold out the ground truth and only release the image samples and questions. We do not provide a development split, as our tasks do not require any specialized knowledge or skills, and we expect decent foundational vision models to solve these tasks without finetuning.
Instead, we provide an evaluation server that scores generated answers, and we maintain an opt-in leaderboard of those.
Evaluations are made by submitting a JSON file with model predictions to our public evaluation server. The server applies our extraction heuristics as outlined above, but users are free to apply their own preprocessing of any kind before submitting their predictions. We rate-limit the server per user and day to prevent ground-truth extraction attacks. 

\subsection{Comparison With Existing Benchmarks}\label{subsec:bench:comparison}

Many existing VQA datasets underestimate the true difficulty of visual reasoning and may probe only for shallow pattern matching rather than genuine scene understanding. Furthermore, they often rely on low-resolution images, recycled content, and automatically generated questions. Our benchmark is intentionally designed to correct these shortcomings and to set a higher standard for evaluation. Its distinguishing features are:
\begin{itemize}%
    \item \textbf{Dense scenes in high-resolution images.} We collect detailed images of complex scenes, enabling questions that demand fine-grained perception and long-range reasoning. Unlike prior benchmarks, which often reduce vision to global features, our dataset forces models to engage with the full richness of the scene--naturally, this is correlated with high-resolution. 
    \item \textbf{Manual annotation.} All questions are crafted by human annotators. Automated pipelines used in other datasets may scale cheaply, but they also introduce biases, trivial patterns, and low-quality queries. Our human-centered approach ensures natural, challenging, and unbiased evaluation.  
    \item \textbf{Fresh image data.} Rather than recycling existing dataset sources, we provide entirely new images. This prevents leakage from pretraining corpora and eliminates the domain biases that plague benchmarks built from reused datasets.  
    \item \textbf{Public domain licensing.} Every image is sourced from the public domain, removing legal barriers that limit distribution or usage. Unlike benchmarks with restrictive or unclear licensing due to web crawling, ours is openly and universally accessible.  
\end{itemize}

In sum, where existing benchmarks compromise on difficulty, reliability, or ethics, our dataset sets a new bar: more challenging, more trustworthy, and more responsible.   

\section{Experiments}\label{sec:experiments}

In the following subsections, we evaluate the performance of different \vlms~on \ourDS. In \cref{subsec:experiments:baselines} we introduce the models, and assess their performance in \cref{subsec:experiments:main}.

\subsection{Baselines}\label{subsec:experiments:baselines}

We evaluate 37 recent \vlms, including variously sized open-weight models ranging from 450M to 109B parameters, designed for low- and high-resolution image understanding, that we separate into three parameter bands, specialized high-resolution understanding models, and 4 proprietary frontier models.
To simplify the answer extraction, we add small postfixes to the benchmark questions outlined in \cref{subsec:appendix:prompts}. We generate answers using greedy decoding for all models, except for proprietary models and models where greedy decoding failed to generate useful outputs (\eg, Llama 4), as highlighted in the result tables.

Additionally, we compare the results to \textit{random chance} (we assume no priors for counting and OCR), as well as \textit{consistent chance}, where we assume that a model is guessing, but gives consistent guesses for logically opposite questions.

\subsection{Main Results}\label{subsec:experiments:main}

The results in \cref{table:main_results} show vast differences between models and some of the tasks in \ourDS. First off, we notice that all models struggle with our freeform counting and OCR tasks. The best accuracy in counting is achieved by Gemini 2.0 Flash, but is only at $41.7 \%$. OCR performance is overall better, but even the best model, o4-mini, only achieves $62.7 \%$. This is also the task with the highest discrepancy between proprietary commercial models and open-weight ones. 

For activity and attribute recognition, we see an improved accuracy (albeit at a higher random chance), but still far from satisfactory performance, even with the strongest models. For reasoning tasks, we find that almost all models struggle and make rather small improvements compared to the consistent random chance, while some of the smaller models even underperform it. The only positive outlier here is o3, which achieves a significant advantage compared to other models, presumably due to its reasoning mode. Unsurprisingly, we find that frontier models achieve a high accuracy on scene understanding, as it primarily relies on a superficial understanding of the scenes, as is common in many of the existing VQA datasets. However, rather surprisingly, the task can still be challenging for many other models, even for large ones. Yet, 8B parameters seem already to be sufficient to achieve $93.4 \%$. In a few cases, the accuracy even fell below a consistent chance, suggesting a fallback to shortcut features (see also \cref{sec:error_analysis}).

Averaged over all tasks, the best model (o3) achieves only $19.6 \%$ on the hardest test split, and $69.5 \%$ overall. The strongest open-weight model is InternVL3 38B with $7.2 \%$ and $67.6 \%$, respectively. Interestingly, we found that specialized HD models perform significantly worse than equally sized regular models. We attribute this primarily to the fact that most \vlms~already apply methodologies such as AnyRes \citep{liu2024llavanext} to support high-resolution images and, thus, performance is rather dependent on the backbones and training, therefore showing that modern \vlms~outperform specialized \vlms~built on older backbones\footnote{Please find an ablation of performance under different resolutions of \ourDS~in \cref{appendix:resolution}.}. Finally, we also find some counter-intuitive scaling trends, where performance decreases with parameter size (often for the largest model of the family, \ie, in InternVL3 and PaliGemma 2).
\begin{table*}
\centering
\caption{\textbf{Benchmark results.} We report the accuracy as the fraction of correct responses after processing, including the accuracy normalization for binary questions for each of the categories in our benchmark, as well as the average. For counting and OCR, we additionally report the RMSE and Normalized Levenshtein Edit Distance, respectively. \textbf{Legend:} $^{S}$ Completions were generated using stochastic sampling at default parameters.}
\label{table:main_results}
\resizebox{\linewidth}{!}{
\renewcommand{\arraystretch}{1.15} 
\begin{tabular}{@{}lccccccc|ccc|c@{}} 
\toprule
 & \textbf{Params}                          & \textbf{Activity} & \textbf{Attributes} & \textbf{Counting} & \textbf{OCR} & \textbf{Reasoning} & \textbf{Scene} & \textbf{Easy} & \textbf{Medium} & \textbf{Hard} & \textbf{Total}  \\
\textbf{Model} & \textbf{[B]} & (150) & (149) & (559) & (118) & (356) & (1388) & (986) & (1304) & (430) & (2720)\\
\midrule
\textcolor{gray}{\textit{Random Chance}}  & - & \textcolor{gray}{25.0} & \textcolor{gray}{25.0} & \textcolor{gray}{0.0} & \textcolor{gray}{0.0} & \textcolor{gray}{25.0} & \textcolor{gray}{25.0} & \textcolor{gray}{24.5} & \textcolor{gray}{16.7} & \textcolor{gray}{3.7} & \textcolor{gray}{16.0}\\
\textcolor{gray}{\textit{Consistent Chance}} & - & \textcolor{gray}{25.0} & \textcolor{gray}{25.0} & \textcolor{gray}{0.0} & \textcolor{gray}{0.0} & \textcolor{gray}{42.5} & \textcolor{gray}{50.0} & \textcolor{gray}{47.2} & \textcolor{gray}{26.2} & \textcolor{gray}{4.7} & \textcolor{gray}{27.2}\\

\midrule
\multicolumn{12}{c}{\textit{Small Open-Weight Models (< 7B)}} \\
\midrule

PaliGemma 2 3B \citep{steiner2024paligemma2} & 3.0 & 42.0 & 53.0 & 20.4 (8.5) & 8.5 (0.69) & 24.9 & 32.7 & 51.9 & 28.3 & 5.0 & 29.0 \\
LLaVA 1.5 7B \citep{llava1_5} & 7.0 & 35.3 & 43.6 & 13.2 (8.2) & 3.4 (0.76) & 39.5 & 43.2 & 69.7 & 24.6 & 1.9 & 30.8 \\
Gemma 3n E2B \citep{Gonzalez2025Gemma3n} & 5.0 & 32.0 & 26.2 & 15.0 (12.2) & 19.5 (0.56) & 35.6 & 53.2 & 74.6 & 25.7 & 7.9 & 33.9 \\
LLaVA-NeXT 7B \citep{liu2024llavanext}& 7.0 & 44.7 & 41.6 & 19.1 (8.6) & 8.5 (0.66) & 40.5 & 54.0 & 81.8 & 31.5 & 2.2 & 37.5 \\
LFM2 VL 450M \citep{liquid2025lfm2} & 0.4 & 35.3 & 47.0 & 22.9 (7.1) & 20.3 (0.57) & 27.8 & 59.5 & 83.1 & 32.4 & 8.6 & 39.7 \\
DeepSeek VL2 Tiny \citep{wu2024deepseekvl2} & 1.0 & 54.7 & 47.7 & 22.5 (8.4) & 35.6 (0.43) & 37.1 & 54.2 & 82.5 & 38.0 & 2.6 & 41.2 \\
SmolVLM \citep{marafioti2025smolvlm} & 2.0 & 42.7 & 41.6 & 17.2 (7.1) & 28.0 (0.47) & 32.2 & 67.3 & 83.5 & 38.8 & 3.1 & 42.0 \\
Gemma 3n E4B \citep{Gonzalez2025Gemma3n} & 5.0 & 40.0 & 23.5 & 19.3 (8.1) & 23.7 (0.54) & 41.0 & 73.9 & 87.8 & 38.4 & 8.9 & 44.2 \\
InternVL3 1B \citep{zhu2025internvl3} & 1.0 & 48.0 & 57.0 & 27.2 (6.2) & 25.4 (0.52) & 35.1 & 77.5 & 94.9 & 48.9 & 5.0 & 50.6 \\
LFM2 VL 1.6B \citep{liquid2025lfm2}& 1.6 & 49.3 & 55.7 & 25.2 (6.2) & 28.0 (0.51) & 44.4 & 79.5 & 97.4 & 50.4 & 4.8 & 51.9 \\
Qwen2.5-VL 3B \citep{bai2025qwen2} & 3.0 & 60.7 & 61.7 & 25.9 (6.1) & 49.2 (0.33) & 43.9 & 77.5 & 94.0 & 56.0 & 4.8 & 54.1 \\
InternVL3 2B \citep{zhu2025internvl3} & 2.0 & 50.0 & 58.4 & 30.4 (8.0) & 39.0 (0.43) & 49.8 & 80.3 & 98.9 & 55.6 & 5.7 & 55.3 \\
DeepSeek VL2 \citep{wu2024deepseekvl2} & 4.5 & 65.3 & 63.8 & 25.9 (8.2) & 46.6 (0.35) & 58.5 & 81.8 & 99.4 & 60.6 & 4.1 & 57.7 \\

\midrule
\multicolumn{12}{c}{\textit{Medium Open-Weight Models (7--13B)}} \\
\midrule

LLaVA-OV 7B \citep{llava-ov} & 7.0 & 60.7 & 57.7 & 28.4 (6.1) & 29.7 (0.45) & 54.1 & 88.2 & 95.5 & 63.6 & 4.3 & 58.3 \\
Qwen2.5-VL 7B \citep{bai2025qwen2} & 7.0 & 63.3 & 69.1 & 34.9 (5.5) & 55.9 (0.27) & 49.8 & 85.3 & 97.9 & 66.2 & 9.6 & 61.5 \\
LLaVA 1.5 13B \citep{llava1_5} & 13.0 & 41.3 & 39.6 & 13.8 (8.1) & 3.4 (0.72) & 42.9 & 71.6 & 94.0 & 34.0 & 2.6 & 42.0 \\
LLaVA-NeXT 13B \citep{liu2024llavanext} & 13.0 & 44.0 & 43.6 & 17.0 (8.3) & 6.8 (0.66) & 41.5 & 75.8 & 97.4 & 38.1 & 2.9 & 45.1 \\
Gemma 3 12B \citep{gemmateam2025gemma3} & 12.0 & 48.7 & 42.3 & 16.5 (9.6) & 31.4 (0.47) & 47.8 & 82.7 & 98.3 & 45.6 & 6.2 & 50.0 \\
PaliGemma 2 10B \citep{steiner2024paligemma2} & 10.0 & 48.7 & 52.3 & 23.6 (7.3) & 5.1 (0.75) & 42.4 & 81.8 & 91.9 & 49.5 & 5.7 & 50.3 \\
InternVL3 8B \citep{zhu2025internvl3} & 8.0 & 66.0 & 67.8 & 32.2 (5.6) & 42.4 (0.37) & 59.0 & 93.4 & \textbf{99.6} & 70.8 & 7.9 & 63.9 \\

\midrule
\multicolumn{12}{c}{\textit{Large Open-Weight Models (> 13B)}} \\
\midrule

PaliGemma 2 28B \citep{steiner2024paligemma2} & 28.0 & 40.0 & 49.0 & 17.4 (9.2) & 5.9 (0.73) & 40.0 & 66.1 & 81.2 & 37.7 & 6.0 & 41.5 \\
Gemma 3 27B \citep{gemmateam2025gemma3} & 27.0 & 51.3 & 46.3 & 18.1 (9.1) & 40.7 (0.41) & 50.7 & 86.3 & 98.5 & 50.6 & 8.9 & 53.2 \\
Llama 4 Scout$^{S}$ \citep{meta2025llama4} & 109.0 & 58.7 & 65.8 & 31.1 (4.2) & 37.3 (0.44) & 62.0 & 78.8 & 95.7 & 57.9 & 13.6 & 57.5 \\
InternVL3 14B \citep{zhu2025internvl3} & 14.0 & 66.7 & 69.1 & 30.6 (5.1) & 41.5 (0.42) & 57.1 & 91.1 & 98.5 & 69.7 & 5.3 & 62.5 \\
LLaVA-OV 72B \citep{llava-ov} & 72.0 & 66.0 & 69.8 & 30.9 (6.0) & 39.0 (0.41) & 57.1 & 91.8 & 97.6 & 71.0 & 4.1 & 62.7 \\
Qwen2.5-VL 32B \citep{bai2025qwen2} & 32.0 & 60.0 & 70.5 & 30.8 (4.7) & 61.0 (0.23) & 61.5 & 90.3 & 98.5 & 68.7 & 12.4 & 63.6 \\
Qwen2.5-VL 72B \citep{bai2025qwen2} & 72.0 & 67.3 & 74.5 & 35.1 (5.7) & 72.9 (0.16) & 53.2 & 90.5 & 97.6 & 72.6 & 13.4 & 65.7 \\
InternVL3 78B \citep{zhu2025internvl3} & 78.0 & \textbf{78.0} & \textbf{80.5} & 34.7 (5.0) & 31.4 (0.52) & 65.4 & 93.7 & 97.6 & 76.9 & 8.1 & 66.8 \\
InternVL3 38B \citep{zhu2025internvl3} & 38.0 & 76.7 & 78.5 & 35.4 (5.3) & 45.8 (0.33) & 69.8 & 92.2 & 98.3 & \textbf{78.6} & 7.2 & 67.6 \\

\midrule
\multicolumn{12}{c}{\textit{Specialized High-Resolution Models}} \\
\midrule

VILA HD 4K$^{S}$ \citep{shi2025scaling} & 8.0 & 54.0 & 48.3 & 22.5 (6.3) & 11.0 (0.73) & 49.3 & 74.5 & 91.2 & 47.1 & 4.1 & 48.5 \\
VILA HD 1.5K$^{S}$ \citep{shi2025scaling} & 8.0 & 54.0 & 57.7 & 25.9 (6.5) & 21.2 (0.52) & 52.2 & 79.4 & 94.2 & 54.3 & 4.1 & 53.1 \\
ILM-XC2-4KHD \citep{ilmxc24khd} & 7.0 & 50.7 & 53.7 & 25.4 (5.8) & 31.4 (0.46) & 42.4 & 83.6 & 94.4 & 53.8 & 6.7 & 53.4 \\
ILM-XC2.5 \citep{zhang2024internlmxcomposer25} & 7.0 & 48.0 & 51.7 & 22.7 (6.0) & 35.6 (0.46) & 45.9 & 87.3 & 95.9 & 53.7 & 9.1 & 54.3 \\

\midrule
\multicolumn{12}{c}{\textit{Proprietary Models}} \\
\midrule
Horizon Alpha$^{S}$ \citep{horizon2025horizonalpha} & -- & 57.3 & 74.5 & 35.6 (5.4) & 48.3 (0.36) & 63.9 & 93.2 & 99.4 & 72.9 & 10.8 & 65.7 \\
Gemini 2.0 Flash & -- & 76.0 & 71.1 & 41.7 (5.8) & 57.6 (0.26) & 56.6 & 92.1 & 99.1 & 74.0 & 19.1 & 68.1 \\
o4-mini$^{S}$ \citep{openai2025o3o4mini} & -- & 70.0 & 76.5 & \textbf{38.3} (4.3) & \textbf{62.7 (0.22)} & 67.8 & 93.7 & 98.1 & 77.4 & 17.2 & 69.1 \\
o3$^{S}$ \citep{openai2025o3o4mini} & -- & 74.0 & 69.8 & 36.7 \textbf{(4.0)} & 61.0 (0.24) & \textbf{75.1} & \textbf{94.7} & 99.4 & 76.4 & \textbf{19.6} & \textbf{69.5} \\
\bottomrule
\end{tabular}
}
\end{table*}

We encourage the community to explore advanced prompting techniques and invite them to submit these to our leaderboard.

\section{Error Analysis}\label{sec:error_analysis}

In this section, we aim to better outline the errors that models make. With the protection of our private ground-truth in mind, we will rely on average statistics over all models described in \cref{subsec:experiments:baselines}.

\begin{figure}
    \centering
    \includegraphics[width=\linewidth]{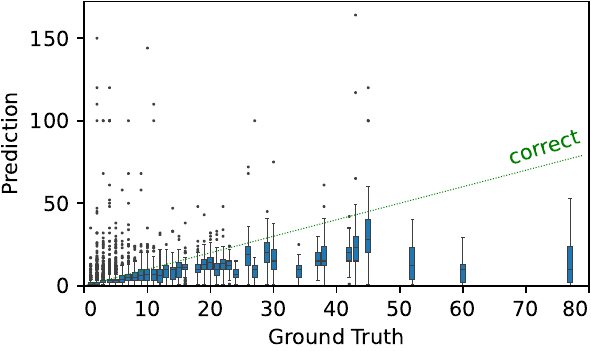}
    \caption{\textbf{Counting prediction vs. ground truths.}}
    \label{fig:counting_error}
\end{figure}

\paragraphx{Counting.} To analyze errors in counting tasks, we plot the distribution of predictions versus ground truths in \cref{fig:counting_error}. Models are generally accurate when the ground truth is low, but errors increase substantially as the ground truth rises. Although some errors stem from incorrect predictions, many are also due to refusals (which we treat as 0) or blank responses (\eg, ``too many objects to count’’). In all cases, models tend to err on the low side and underestimate the ground truth. Yet, our analysis also contained outliers showing severe overestimation.

\begin{figure}
    \centering
    \includegraphics[width=\linewidth]{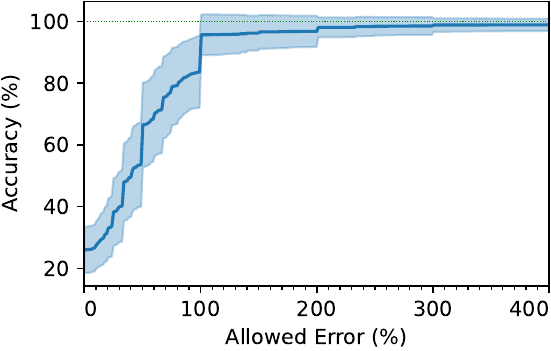}
    \caption{\textbf{Counting accuracy under tolerance} (mean$\pm$std).}
    \label{fig:counting_tolerance}
\end{figure}

To quantify the magnitude of these errors, we measured accuracy under varying tolerance levels, shown in \cref{fig:counting_tolerance}. Prediction errors are typically severe: even with a $10 \%$ tolerance, average accuracy improves by only $1.6 \%$. Larger tolerances, such as $50 \%$ or $100 \%$, yield more substantial improvements, but such levels are impractical for real-world applications.

\begin{figure}
    \centering
    \includegraphics[width=\linewidth]{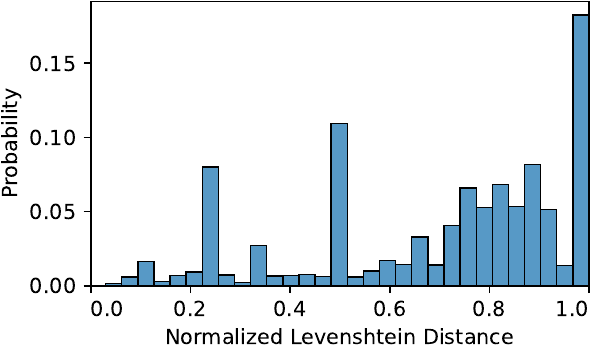}
    \caption{\textbf{OCR prediction error distance.}}
    \label{fig:ocr_distance}
\end{figure}

\paragraphx{OCR.} Similar to counting, we aim to quantify the magnitude of errors in OCR predictions. To do this, we measure the Levenshtein edit distance \citep{Lev65} between preprocessed predictions (as described in \cref{subsec:bench:eval}) and ground truths for incorrect answers. We normalize the distance by the maximum sequence length and visualize the distribution in \cref{fig:ocr_distance}. The distribution’s center of mass is around $0.7$, indicating that sequences require substantial edits to be correct, highlighting severe errors.

Manual inspection of a subset of errors reveals three main causes: hallucinations, extraction of irrelevant text, and, in a few severe cases, failure to follow the instruction to respond only with the text. Errors of the second type often arise from misinterpretation of text flow, such as side-by-side multi-line paragraphs or non-standard layouts like banners. For errors with low edit distance, we frequently observe that models' auto-correct spelling or generally fall back to more probable token sequences rather than reproducing the actual text (\eg, ``accidunt'' becomes ``accident''), particularly in non-English or non-Latin scripts.

\begin{figure}%
    \centering
    \includegraphics[width=\linewidth]{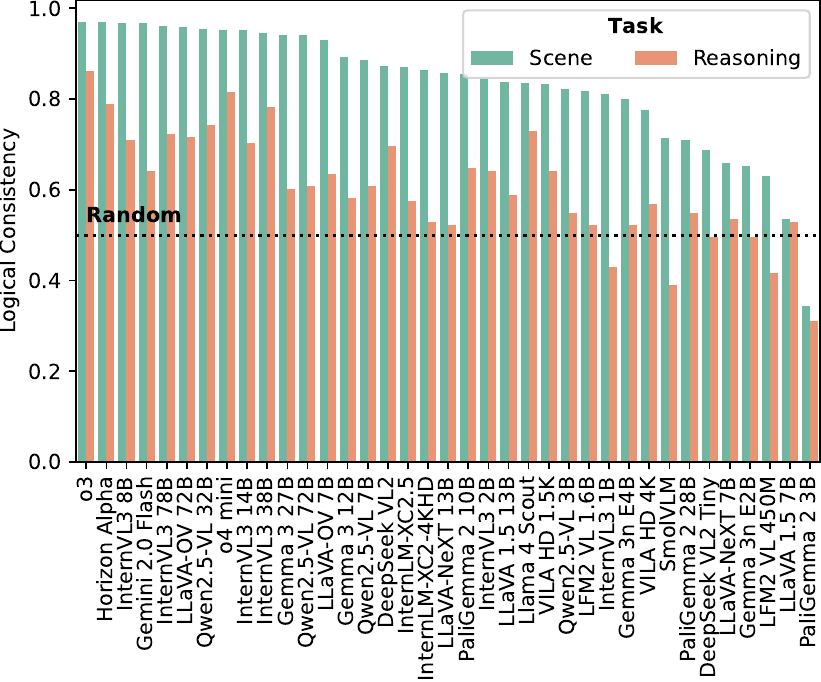}
    \caption{\textbf{Logical consistency on \textit{scene understanding} and \textit{reasoning}.}}
    \label{fig:logical_consistency}
\end{figure}

\paragraphx{Logical consistency.} As described in \cref{subsec:bench:curation}, our dataset contains binary questions, where each such question is paired with a logical opposite. A strong model should argue logically consistently, even if the answer is wrong. For instance, if a model answers ``yes'' to ``Is it day?'' it should answer ``no'' to ``Is it night?''. We measure the ratio of logically consistent answer pairs per model and task (reasoning and global scene understanding) and visualize the results in \cref{fig:logical_consistency}. 

We observe that frontier models answer fairly logically consistent for the easier scene questions, but their performance rapidly drops on the harder reasoning questions. On average, consistency falls from $83.3 \%$ or $60.6 \%$ between the tasks. For some models, a well-above-chance consistency drops a near-random baseline for reasoning, suggesting that models are now guessing independently of the context, while providing well-grounded answers on the original task. In some cases, we also find a well-below random chance consistency, suggesting that the model is relying on shortcuts for shortcuts rather than the visual inputs. Alarmingly, we find PaliGemma2 3B to be susceptible to these for both tasks. 

\section{Related Work}\label{sec:relatedwork}

\paragraphx{Large multi-modal models.} 
Recent progress in \vlms~has significantly advanced the integration of visual and linguistic modalities, 
enabling more sophisticated multi-modal understanding and generation. 
Early approaches connect pretrained vision encoders with large language models via lightweight modules, achieving competitive performance with relatively few trainable parameters \citep{li2023blip,zhang2023llama,minigpt4}.
The LLaVA series~\citep{llava,llava1_5,liu2024llavanext,llava-ov} improves visual instruction tuning, demonstrating stronger performance on fine-grained visual tasks.
More recent models extend these capabilities to multi-image contexts, enabling richer scene understanding and more coherent textual reasoning \citep{llava-ov,steiner2024paligemma2,zhu2025internvl3,bai2025qwen2,gemmateam2025gemma3}.
Proprietary \vlms, including GPT, the o-series \citep{openai2024gpt4,openai2025o3o4mini}, and Gemini \citep{geminiteam2024gemini,google2025gemini2flash}, further highlight progress in versatile, context-aware multi-modal learning frameworks, sometimes even including multi-modal reasoning \citep{openai2025o3o4mini}.

Despite these advancements, many VLMs still exhibit notable weaknesses in visual understanding \citep{fu2025blink,vo2025visionlanguagemodelsbiased}. %
Prior work has shown that they struggle with counting, spatial reasoning, concept binding, and dense scene understanding \citep{Paiss_2023_ICCV,yin2023lamm,xu2025lvlmhum,doveh2023dense, doveh2023teaching, huang2024conme, campbell2024understanding} and even image classification tasks \citep{mirza2024glov, mirza2023lafter, zhang2024visually}.
In our work, we build on these findings by introducing a benchmark of densely populated public-domain paintings, designed to probe such vulnerabilities and evaluate the capacity of VLMs to perform basic visual tasks in more challenging, \textit{visually overloaded} scenes.

\paragraphx{Multi-modal vision benchmarks.}
The rapid progress of \vlms~has spurred a surge of benchmarks evaluating their ability to integrate vision and language across tasks such as VQA, captioning, reasoning, and instruction following. Extending classic VQA datasets \citep{antol2015vqa,Goyal_2017_CVPR}, modern benchmarks vary in scope, from real-world instruction following in VisitBench \citep{visitbench2023} to conversational reasoning in LLaVA-Bench \citep{llavabench2023}, zero-shot capability assessment across 16 capabilities, including OCR and spatial reasoning in MMVet \citep{mmvet2024}, and multiple-choice probing in 12 dimensions in SeedBench \citep{seedbench2023}. Broader frameworks such as MM-Bench \citep{mmbench2023}, TouchStone \citep{touchstone2024}, OmniBench \citep{omnibench2025}, and MMStar \citep{mmstar2024} aim for holistic multi-modal evaluation by covering a wide array of tasks and domain-specific knowledge.
MMMU \citep{yue2024mmmu} pushes toward expert-level multi-modal reasoning. 
As performance on most of these benchmarks seems to saturate, more carefully designed benchmarks \citep{qbench2023,huang2024conme,thrush2022winoground,hsieh2023sugarcrepe} reveal persistent weaknesses in multiple dimensions, highlighting a discrepancy between many seemingly positive benchmark results and actual visual capabilities.

While these efforts nonetheless provide valuable insights, most emphasize global understanding, a very broad task coverage, or require domain-specific expertise, while often overlooking basic perception in more challenging settings, such as visually overloaded scenes. Recently, multiple benchmarks started the exploration of small details in high-resolution scenes \citep{vstar2023,otterhd,shi2025scaling} or documents \citep{Mathew_2021_WACV,Tanaka2023slidevqa}, showing another hurdle in the development of vision models. Our work complements these benchmarks with \ourDS, a human-annotated dataset of VQA pairs grounded in high-resolution, but also densely populated artworks. A key differentiator of high-resolution benchmarks is that \ourDS~aims at exploiting the full complexity of the scene, while previous works mostly model needle-in-the-haystack-style retrieval of small details. By focusing on six basic tasks in overloaded scenes, \ourDS~reveals systematic error modes in state-of-the-art open and proprietary \vlms, highlighting critical gaps even in basic, knowledge-free visual understanding.

\section{Conclusion}\label{sec:conclusion}
In this work, we introduced \ourDS, a novel VQA benchmark designed to expose the limitations of state-of-the-art \vlms~in complex, detail-rich scenes. Our findings demonstrate that while these models perform well on global tasks, they consistently struggle with simple, fine-grained questions within \textit{visually overloaded} environments. This performance gap highlights a critical area for future research, suggesting that the problem of fundamental visual understanding is far from solved. Ultimately, our dataset offers a crucial resource for the community to develop more robust and perceptive \vlms.

\paragraph{Limitations.} Our findings are based on a dataset composed exclusively of art, which means they may not directly generalize to benchmarks of natural images. Artistic representations often diverge significantly from photorealism, emphasizing abstractness over detail \cite{tversky2002sketches}. This introduces a unique set of challenges: while models trained on natural images may excel at texture-based recognition \citep{geirhos2018imagenettrained,gavrikov2025can}, they often struggle with the high stylistic variance and strong reliance on global shape inherent in art. We therefore consider this focus a feature, not a drawback. We posit that a \textit{truly} foundational model, aiming to match or exceed human-level perception, must also be robust to these variations and comprehend representations just as humans do. We further elaborate on this in \cref{subsec:appendix:domainshift}.

\clearpage
    
{
    \small
    \bibliographystyle{ieeenat_fullname}
    \bibliography{main}

\begin{thebibliography}{76}
\providecommand{\natexlab}[1]{#1}
\providecommand{\url}[1]{\texttt{#1}}
\expandafter\ifx\csname urlstyle\endcsname\relax
  \providecommand{\doi}[1]{doi: #1}\else
  \providecommand{\doi}{doi: \begingroup \urlstyle{rm}\Url}\fi

\bibitem[Agrawal et~al.(2016)Agrawal, Batra, and Parikh]{agrawal2016analyzing}
Aishwarya Agrawal, Dhruv Batra, and Devi Parikh.
\newblock {Analyzing the Behavior of Visual Question Answering Models}.
\newblock \emph{arXiv preprint arXiv:1606.07356}, 2016.

\bibitem[Agrawal et~al.(2018)Agrawal, Batra, Parikh, and Kembhavi]{Agrawal_2018_CVPR}
Aishwarya Agrawal, Dhruv Batra, Devi Parikh, and Aniruddha Kembhavi.
\newblock {Don't Just Assume; Look and Answer: Overcoming Priors for Visual Question Answering}.
\newblock In \emph{Proceedings of the Conference on Computer Vision and Pattern Recognition (CVPR)}, 2018.

\bibitem[Antol et~al.(2015)Antol, Agrawal, Lu, Mitchell, Batra, Zitnick, and Parikh]{antol2015vqa}
Stanislaw Antol, Aishwarya Agrawal, Jiasen Lu, Margaret Mitchell, Dhruv Batra, C.~Lawrence Zitnick, and Devi Parikh.
\newblock {{VQA}: {V}isual {Q}uestion {A}nswering}.
\newblock In \emph{Proceedings of the International Conference on Computer Vision (ICCV)}, 2015.

\bibitem[Bai et~al.(2023)Bai, Yang, Bai, Wang, Zhang, Lin, Wang, Zhou, and Zhou]{touchstone2024}
Shuai Bai, Shusheng Yang, Jinze Bai, Peng Wang, Xingxuan Zhang, Junyang Lin, Xinggang Wang, Chang Zhou, and Jingren Zhou.
\newblock {Touchstone: Evaluating vision-language models by language models}.
\newblock \emph{arXiv preprint arXiv:2308.16890}, 2023.

\bibitem[Bai et~al.(2025)Bai, Chen, Liu, Wang, Ge, Song, Dang, Wang, Wang, Tang, et~al.]{bai2025qwen2}
Shuai Bai, Keqin Chen, Xuejing Liu, Jialin Wang, Wenbin Ge, Sibo Song, Kai Dang, Peng Wang, Shijie Wang, Jun Tang, et~al.
\newblock {Qwen2.5-VL technical report}.
\newblock \emph{arXiv preprint arXiv:2502.13923}, 2025.

\bibitem[Becattini et~al.(2023)Becattini, Bongini, Bulla, Bimbo, Marinucci, Mongiov\`{\i}, and Presutti]{Becattini2023VISCOUNTH}
Federico Becattini, Pietro Bongini, Luana Bulla, Alberto~Del Bimbo, Ludovica Marinucci, Misael Mongiov\`{\i}, and Valentina Presutti.
\newblock {VISCOUNTH: A Large-scale Multilingual Visual Question Answering Dataset for Cultural Heritage}.
\newblock \emph{ACM Trans. Multimedia Comput. Commun. Appl.}, 19\penalty0 (6), 2023.

\bibitem[Bitton et~al.(2023)Bitton, Bansal, Hessel, Shao, Zhu, Awadalla, Gardner, Taori, and Schmidt]{visitbench2023}
Yonatan Bitton, Hritik Bansal, Jack Hessel, Rulin Shao, Wanrong Zhu, Anas Awadalla, Josh Gardner, Rohan Taori, and Ludwig Schmidt.
\newblock {Visit-bench: A benchmark for vision-language instruction following inspired by real-world use}.
\newblock \emph{arXiv preprint arXiv:2308.06595}, 2023.

\bibitem[Bordes et~al.(2024)Bordes, Pang, Ajay, Li, Bardes, Petryk, Ma{\~n}as, Lin, Mahmoud, Jayaraman, et~al.]{llavabench2023}
Florian Bordes, Richard~Yuanzhe Pang, Anurag Ajay, Alexander~C Li, Adrien Bardes, Suzanne Petryk, Oscar Ma{\~n}as, Zhiqiu Lin, Anas Mahmoud, Bargav Jayaraman, et~al.
\newblock {An introduction to vision-language modeling}.
\newblock \emph{arXiv preprint arXiv:2405.17247}, 2024.

\bibitem[Cadene et~al.(2019)Cadene, Dancette, Ben~younes, Cord, and Parikh]{cadene2019rubi}
Remi Cadene, Corentin Dancette, Hedi Ben~younes, Matthieu Cord, and Devi Parikh.
\newblock {RUBi: Reducing Unimodal Biases for Visual Question Answering}.
\newblock In \emph{Advances in Neural Information Processing Systems (NeurIPS)}, 2019.

\bibitem[Campbell et~al.(2024)Campbell, Rane, Giallanza, De~Sabbata, Ghods, Joshi, Ku, Frankland, Griffiths, Cohen, and Webb]{campbell2024understanding}
Declan Campbell, Sunayana Rane, Tyler Giallanza, Nicol\`{o} De~Sabbata, Kia Ghods, Amogh Joshi, Alexander Ku, Steven~M. Frankland, Thomas~L. Griffiths, Jonathan~D. Cohen, and Taylor Webb.
\newblock {Understanding the Limits of Vision Language Models Through the Lens of the Binding Problem}.
\newblock In \emph{Advances in Neural Information Processing Systems (NeurIPS)}, 2024.

\bibitem[Chen et~al.(2024)Chen, Li, Dong, Zhang, Zang, Chen, Duan, Wang, Qiao, Lin, et~al.]{mmstar2024}
Lin Chen, Jinsong Li, Xiaoyi Dong, Pan Zhang, Yuhang Zang, Zehui Chen, Haodong Duan, Jiaqi Wang, Yu Qiao, Dahua Lin, et~al.
\newblock {Are we on the right way for evaluating large vision-language models?}
\newblock \emph{Advances in Neural Information Processing Systems (NeurIPS)}, 37, 2024.

\bibitem[Dong et~al.(2024)Dong, Zhang, Zang, Cao, Wang, Ouyang, Zhang, Duan, Zhang, Li, et~al.]{ilmxc24khd}
Xiaoyi Dong, Pan Zhang, Yuhang Zang, Yuhang Cao, Bin Wang, Linke Ouyang, Songyang Zhang, Haodong Duan, Wenwei Zhang, Yining Li, et~al.
\newblock {Internlm-xcomposer2-4khd: A pioneering large vision-language model handling resolutions from 336 pixels to 4k hd}.
\newblock \emph{Advances in Neural Information Processing Systems (NeurIPS)}, 37, 2024.

\bibitem[Doveh et~al.(2023{\natexlab{a}})Doveh, Arbelle, Harary, Herzig, Kim, Cascante-Bonilla, Alfassy, Panda, Giryes, Feris, et~al.]{doveh2023dense}
Sivan Doveh, Assaf Arbelle, Sivan Harary, Roei Herzig, Donghyun Kim, Paola Cascante-Bonilla, Amit Alfassy, Rameswar Panda, Raja Giryes, Rogerio Feris, et~al.
\newblock {Dense and aligned captions (dac) promote compositional reasoning in vl models}.
\newblock \emph{Advances in Neural Information Processing Systems (NeurIPS)}, 36, 2023{\natexlab{a}}.

\bibitem[Doveh et~al.(2023{\natexlab{b}})Doveh, Arbelle, Harary, Schwartz, Herzig, Giryes, Feris, Panda, Ullman, and Karlinsky]{doveh2023teaching}
Sivan Doveh, Assaf Arbelle, Sivan Harary, Eli Schwartz, Roei Herzig, Raja Giryes, Rogerio Feris, Rameswar Panda, Shimon Ullman, and Leonid Karlinsky.
\newblock {Teaching structured vision \& language concepts to vision \& language models}.
\newblock In \emph{Proceedings of the Conference on Computer Vision and Pattern Recognition (CVPR)}, 2023{\natexlab{b}}.

\bibitem[Forsythe et~al.(2011)Forsythe, Nadal, Sheehy, Cela-Conde, and Sawey]{forsythe2011predicting}
A. Forsythe, M. Nadal, N. Sheehy, C.~J. Cela-Conde, and M. Sawey.
\newblock {Predicting beauty: Fractal dimension and visual complexity in art}.
\newblock \emph{British Journal of Psychology}, 102\penalty0 (1), 2011.

\bibitem[Fu et~al.(2025)Fu, Hu, Li, Feng, Wang, Lin, Roth, Smith, Ma, and Krishna]{fu2025blink}
Xingyu Fu, Yushi Hu, Bangzheng Li, Yu Feng, Haoyu Wang, Xudong Lin, Dan Roth, Noah~A. Smith, Wei-Chiu Ma, and Ranjay Krishna.
\newblock {BLINK: Multimodal Large Language Models Can See but Not Perceive}.
\newblock In \emph{Proceedings of the European Conference on Computer Vision (ECCV)}, 2025.

\bibitem[Garcia et~al.(2020)Garcia, Ye, Liu, Hu, Otani, Chu, Nakashima, and Mitamura]{Garcia2020aqua}
Noa Garcia, Chentao Ye, Zihua Liu, Qingtao Hu, Mayu Otani, Chenhui Chu, Yuta Nakashima, and Teruko Mitamura.
\newblock {A Dataset and Baselines for Visual Question Answering on Art}.
\newblock In \emph{Proceedings of the European Conference on Computer Vision Workshops (ECCVW)}, 2020.

\bibitem[Gavrikov et~al.(2025)Gavrikov, Lukasik, Jung, Geirhos, Mirza, Keuper, and Keuper]{gavrikov2025can}
Paul Gavrikov, Jovita Lukasik, Steffen Jung, Robert Geirhos, Muhammad~Jehanzeb Mirza, Margret Keuper, and Janis Keuper.
\newblock {Can We Talk Models Into Seeing the World Differently?}
\newblock In \emph{International Conference on Learning Representations (ICLR)}, 2025.

\bibitem[Gebru et~al.(2021)Gebru, Morgenstern, Vecchione, Vaughan, Wallach, III, and Crawford]{gebru2021datasheet}
Timnit Gebru, Jamie Morgenstern, Briana Vecchione, Jennifer~Wortman Vaughan, Hanna Wallach, Hal~Daum\'{e} III, and Kate Crawford.
\newblock {Datasheets for datasets}.
\newblock \emph{Commun. ACM}, 64\penalty0 (12), 2021.

\bibitem[Geirhos et~al.(2019)Geirhos, Rubisch, Michaelis, Bethge, Wichmann, and Brendel]{geirhos2018imagenettrained}
Robert Geirhos, Patricia Rubisch, Claudio Michaelis, Matthias Bethge, Felix~A. Wichmann, and Wieland Brendel.
\newblock {ImageNet-trained {CNN}s are biased towards texture; increasing shape bias improves accuracy and robustness.}
\newblock In \emph{International Conference on Learning Representations (ICLR)}, 2019.

\bibitem[{Gemini Team}(2024)]{geminiteam2024gemini}
{Gemini Team}.
\newblock {Gemini: A Family of Highly Capable Multimodal Models}.
\newblock \emph{arXiv preprint arXiv:2312.11805}, 2024.

\bibitem[{Gemini Team}(2025)]{google2025gemini2flash}
{Gemini Team}.
\newblock {Gemini 2.0 Flash Model Card}, 2025.
\newblock [Online; accessed 28. Aug. 2025].

\bibitem[{Gemma Team}(2025)]{gemmateam2025gemma3}
{Gemma Team}.
\newblock {Gemma 3 Technical Report}.
\newblock \emph{arXiv preprint arXiv:2503.19786}, 2025.

\bibitem[Gonzalez and Shivanna(2025)]{Gonzalez2025Gemma3n}
Lucas Gonzalez and Rakesh Shivanna.
\newblock {Announcing Gemma 3n preview: powerful, efficient, mobile-first AI}, 2025.
\newblock [Online; accessed 28. Aug. 2025].

\bibitem[Goyal et~al.(2017)Goyal, Khot, Summers-Stay, Batra, and Parikh]{Goyal_2017_CVPR}
Yash Goyal, Tejas Khot, Douglas Summers-Stay, Dhruv Batra, and Devi Parikh.
\newblock {Making the v in VQA Matter: Elevating the Role of Image Understanding in Visual Question Answering}.
\newblock In \emph{Proceedings of the Conference on Computer Vision and Pattern Recognition (CVPR)}, 2017.

\bibitem[{Horizon Alpha Team}(2025)]{horizon2025horizonalpha}
{Horizon Alpha Team}.
\newblock {Horizon Alpha - Advanced AI Language Model}, 2025.
\newblock [Online; accessed 28. Aug. 2025].

\bibitem[Hsieh et~al.(2023)Hsieh, Zhang, Ma, Kembhavi, and Krishna]{hsieh2023sugarcrepe}
Cheng-Yu Hsieh, Jieyu Zhang, Zixian Ma, Aniruddha Kembhavi, and Ranjay Krishna.
\newblock {Sugarcrepe: Fixing hackable benchmarks for vision-language compositionality}.
\newblock \emph{Advances in Neural Information Processing Systems (NeurIPS)}, 36, 2023.

\bibitem[Huang et~al.(2024)Huang, Lin, Mirza, Hansen, Doveh, Butoi, Herzig, Arbelle, Kuehne, Darrell, et~al.]{huang2024conme}
Irene Huang, Wei Lin, Muhammad~Jehanzeb Mirza, Jacob Hansen, Sivan Doveh, Victor Butoi, Roei Herzig, Assaf Arbelle, Hilde Kuehne, Trevor Darrell, et~al.
\newblock {Conme: Rethinking evaluation of compositional reasoning for modern vlms}.
\newblock \emph{Advances in Neural Information Processing Systems (NeurIPS)}, 37, 2024.

\bibitem[Kojima et~al.(2022)Kojima, Gu, Reid, Matsuo, and Iwasawa]{kojima2022cot}
Takeshi Kojima, Shixiang~(Shane) Gu, Machel Reid, Yutaka Matsuo, and Yusuke Iwasawa.
\newblock {Large Language Models are Zero-Shot Reasoners}.
\newblock In \emph{Advances in Neural Information Processing Systems (NeurIPS)}, 2022.

\bibitem[Levenshtein(1965)]{Lev65}
Vladimir~Iosifovich Levenshtein.
\newblock {Dvoichnye kody s ispravleniem vypadenii, vstavok i zameshchenii simvolov}.
\newblock \emph{Doklady Akademii Nauk SSSR}, 163\penalty0 (4), 1965.

\bibitem[Li et~al.(2023{\natexlab{a}})Li, Wang, Wang, Ge, Ge, and Shan]{seedbench2023}
Bohao Li, Rui Wang, Guangzhi Wang, Yuying Ge, Yixiao Ge, and Ying Shan.
\newblock {Seed-bench: Benchmarking multimodal llms with generative comprehension}.
\newblock \emph{arXiv preprint arXiv:2307.16125}, 2023{\natexlab{a}}.

\bibitem[Li et~al.(2023{\natexlab{b}})Li, Zhang, Yang, Zhang, Pu, and Liu]{otterhd}
Bo Li, Peiyuan Zhang, Jingkang Yang, Yuanhan Zhang, Fanyi Pu, and Ziwei Liu.
\newblock {Otterhd: A high-resolution multi-modality model}.
\newblock \emph{arXiv preprint arXiv:2311.04219}, 2023{\natexlab{b}}.

\bibitem[Li et~al.(2024{\natexlab{a}})Li, Zhang, Guo, Zhang, Li, Zhang, Zhang, Li, Liu, and Li]{llava-ov}
Bo Li, Yuanhan Zhang, Dong Guo, Renrui Zhang, Feng Li, Hao Zhang, Kaichen Zhang, Yanwei Li, Ziwei Liu, and Chunyuan Li.
\newblock {LLaVA-OneVision: Easy Visual Task Transfer}.
\newblock \emph{arXiv preprint arXiv:2408.03326}, 2024{\natexlab{a}}.

\bibitem[Li et~al.(2023{\natexlab{c}})Li, Li, Savarese, and Hoi]{li2023blip}
Junnan Li, Dongxu Li, Silvio Savarese, and Steven Hoi.
\newblock {Blip-2: Bootstrapping language-image pre-training with frozen image encoders and large language models}.
\newblock In \emph{Proceedings of the International Conference on Machine Learning (ICML)}, 2023{\natexlab{c}}.

\bibitem[Li et~al.(2024{\natexlab{b}})Li, Zhang, Ma, Yuan, Zhu, Guo, Liang, Liu, Wang, Yang, et~al.]{omnibench2025}
Yizhi Li, Ge Zhang, Yinghao Ma, Ruibin Yuan, Kang Zhu, Hangyu Guo, Yiming Liang, Jiaheng Liu, Zekun Wang, Jian Yang, et~al.
\newblock {Omnibench: Towards the future of universal omni-language models}.
\newblock \emph{arXiv preprint arXiv:2409.15272}, 2024{\natexlab{b}}.

\bibitem[Lin et~al.(2025)Lin, Ru, Rapp, Guan, and Xiong~Bearfield]{lin2025what}
Kylie Lin, Sean Sheng-tse Ru, David~N. Rapp, Hui Guan, and Cindy Xiong~Bearfield.
\newblock {What Makes a Visualization Visually Complex?}
\newblock In \emph{Proceedings of the Extended Abstracts of the CHI Conference on Human Factors in Computing Systems}, 2025.

\bibitem[{Liquid AI}(2025)]{liquid2025lfm2}
{Liquid AI}.
\newblock {Introducing LFM2: The Fastest On-Device Foundation Models on the Market {$\vert$} Liquid AI}, 2025.
\newblock [Online; accessed 28. Aug. 2025].

\bibitem[Liu et~al.(2023{\natexlab{a}})Liu, Li, Li, and Lee]{llava1_5}
Haotian Liu, Chunyuan Li, Yuheng Li, and Yong~Jae Lee.
\newblock {Improved baselines with visual instruction tuning}.
\newblock \emph{arXiv preprint arXiv:2310.03744}, 2023{\natexlab{a}}.

\bibitem[Liu et~al.(2023{\natexlab{b}})Liu, Li, Wu, and Lee]{llava}
Haotian Liu, Chunyuan Li, Qingyang Wu, and Yong~Jae Lee.
\newblock {Visual instruction tuning}.
\newblock \emph{Advances in Neural Information Processing Systems (NeurIPS)}, 36, 2023{\natexlab{b}}.

\bibitem[Liu et~al.(2024{\natexlab{a}})Liu, Li, Li, Li, Zhang, Shen, and Lee]{liu2024llavanext}
Haotian Liu, Chunyuan Li, Yuheng Li, Bo Li, Yuanhan Zhang, Sheng Shen, and Yong~Jae Lee.
\newblock {LLaVA-NeXT: Improved reasoning, OCR, and world knowledge}, 2024{\natexlab{a}}.

\bibitem[Liu et~al.(2024{\natexlab{b}})Liu, Duan, Zhang, Li, Zhang, Zhao, Yuan, Wang, He, Liu, et~al.]{mmbench2023}
Yuan Liu, Haodong Duan, Yuanhan Zhang, Bo Li, Songyang Zhang, Wangbo Zhao, Yike Yuan, Jiaqi Wang, Conghui He, Ziwei Liu, et~al.
\newblock {Mmbench: Is your multi-modal model an all-around player?}
\newblock In \emph{Proceedings of the European Conference on Computer Vision (ECCV)}. Springer, 2024{\natexlab{b}}.

\bibitem[Mahon and Lukasiewicz(2023)]{mahon2023minimumdescriptionlengthclustering}
Louis Mahon and Thomas Lukasiewicz.
\newblock {Minimum Description Length Clustering to Measure Meaningful Image Complexity}.
\newblock \emph{arXiv preprint arXiv:2306.14937}, 2023.

\bibitem[Marafioti et~al.(2025)Marafioti, Zohar, Farré, Noyan, Bakouch, Cuenca, Zakka, Allal, Lozhkov, Tazi, Srivastav, Lochner, Larcher, Morlon, Tunstall, von Werra, and Wolf]{marafioti2025smolvlm}
Andrés Marafioti, Orr Zohar, Miquel Farré, Merve Noyan, Elie Bakouch, Pedro Cuenca, Cyril Zakka, Loubna~Ben Allal, Anton Lozhkov, Nouamane Tazi, Vaibhav Srivastav, Joshua Lochner, Hugo Larcher, Mathieu Morlon, Lewis Tunstall, Leandro von Werra, and Thomas Wolf.
\newblock {SmolVLM: Redefining small and efficient multimodal models}.
\newblock \emph{arXiv preprint arXiv:2504.05299}, 2025.

\bibitem[Marin and Leder(2013)]{marin2014examining}
Manuela~M. Marin and Helmut Leder.
\newblock {Examining Complexity across Domains: Relating Subjective and Objective Measures of Affective Environmental Scenes, Paintings and Music}.
\newblock \emph{PLOS ONE}, 8\penalty0 (8), 2013.

\bibitem[Mathew et~al.(2021)Mathew, Karatzas, and Jawahar]{Mathew_2021_WACV}
Minesh Mathew, Dimosthenis Karatzas, and C.V. Jawahar.
\newblock {DocVQA: A Dataset for VQA on Document Images}.
\newblock In \emph{IEEE Winter Conference on Applications of Computer Vision (WACV)}, 2021.

\bibitem[McInnes et~al.(2018)McInnes, Healy, Saul, and Großberger]{McInnes2018umap}
Leland McInnes, John Healy, Nathaniel Saul, and Lukas Großberger.
\newblock {UMAP: Uniform Manifold Approximation and Projection}.
\newblock \emph{Journal of Open Source Software}, 3\penalty0 (29), 2018.

\bibitem[{Meta AI}(2025)]{meta2025llama4}
{Meta AI}.
\newblock {The Llama 4 herd: The beginning of a new era of natively multimodal AI innovation}, 2025.
\newblock [Online; accessed 28. Aug. 2025].

\bibitem[Mirza et~al.(2023)Mirza, Karlinsky, Lin, Possegger, Kozinski, Feris, and Bischof]{mirza2023lafter}
Muhammad~Jehanzeb Mirza, Leonid Karlinsky, Wei Lin, Horst Possegger, Mateusz Kozinski, Rogerio Feris, and Horst Bischof.
\newblock {LaFTer: Label-Free Tuning of Zero-shot Classifier using Language and Unlabeled Image Collections}.
\newblock \emph{Advances in Neural Information Processing Systems (NeurIPS)}, 36, 2023.

\bibitem[Mirza et~al.(2025)Mirza, Zhao, Mao, Doveh, Lin, Gavrikov, Dorkenwald, Yang, Jha, Wakaki, Mitsufuji, Possegger, Feris, Karlinsky, and Glass]{mirza2024glov}
Muhammad~Jehanzeb Mirza, Mengjie Zhao, Zhuoyuan Mao, Sivan Doveh, Wei Lin, Paul Gavrikov, Michael Dorkenwald, Shiqi Yang, Saurav Jha, Hiromi Wakaki, Yuki Mitsufuji, Horst Possegger, Rogerio Feris, Leonid Karlinsky, and James~R. Glass.
\newblock {{GLOV}: Guided Large Language Models as Implicit Optimizers for Vision Language Models}.
\newblock \emph{Transactions on Machine Learning Research (TMLR)}, 2025.

\bibitem[{OpenAI}(2024)]{openai2024gpt4}
{OpenAI}.
\newblock {GPT-4 Technical Report}.
\newblock \emph{arXiv preprint arXiv:2303.08774}, 2024.

\bibitem[{OpenAI}(2025)]{openai2025o3o4mini}
{OpenAI}.
\newblock {OpenAI o3 and o4-mini System Card}, 2025.
\newblock [Online; accessed 28. Aug. 2025].

\bibitem[Paiss et~al.(2023)Paiss, Ephrat, Tov, Zada, Mosseri, Irani, and Dekel]{Paiss_2023_ICCV}
Roni Paiss, Ariel Ephrat, Omer Tov, Shiran Zada, Inbar Mosseri, Michal Irani, and Tali Dekel.
\newblock {Teaching CLIP to Count to Ten}.
\newblock In \emph{Proceedings of the International Conference on Computer Vision (ICCV)}, 2023.

\bibitem[Phan et~al.(2025)]{phan2025humanitysexam}
Long Phan et~al.
\newblock {Humanity's Last Exam}.
\newblock \emph{arXiv preprint arXiv:2501.14249}, 2025.

\bibitem[Shi et~al.(2025)Shi, Li, Cai, Lu, Liu, Pavone, Kautz, Han, Darrell, Molchanov, and Yin]{shi2025scaling}
Baifeng Shi, Boyi Li, Han Cai, Yao Lu, Sifei Liu, Marco Pavone, Jan Kautz, Song Han, Trevor Darrell, Pavlo Molchanov, and Hongxu Yin.
\newblock {Scaling Vision Pre-Training to 4K Resolution}.
\newblock \emph{arXiv preprint arXiv:2503.19903}, 2025.

\bibitem[Steiner et~al.(2024)Steiner, Pinto, Tschannen, Keysers, Wang, Bitton, Gritsenko, Minderer, Sherbondy, Long, Qin, Ingle, Bugliarello, Kazemzadeh, Mesnard, Alabdulmohsin, Beyer, and Zhai]{steiner2024paligemma2}
Andreas Steiner, André~Susano Pinto, Michael Tschannen, Daniel Keysers, Xiao Wang, Yonatan Bitton, Alexey Gritsenko, Matthias Minderer, Anthony Sherbondy, Shangbang Long, Siyang Qin, Reeve Ingle, Emanuele Bugliarello, Sahar Kazemzadeh, Thomas Mesnard, Ibrahim Alabdulmohsin, Lucas Beyer, and Xiaohua Zhai.
\newblock {PaliGemma 2: A Family of Versatile VLMs for Transfer}.
\newblock \emph{arXiv preprint arXiv:2412.03555}, 2024.

\bibitem[Stim(2023)]{RichardStim2023Jan}
Richard Stim.
\newblock {How Long Does Copyright Protection Last?}
\newblock \emph{Nolo}, 2023.

\bibitem[Tanaka et~al.(2023)Tanaka, Nishida, Nishida, Hasegawa, Saito, and Saito]{Tanaka2023slidevqa}
Ryota Tanaka, Kyosuke Nishida, Kosuke Nishida, Taku Hasegawa, Itsumi Saito, and Kuniko Saito.
\newblock {SlideVQA: a dataset for document visual question answering on multiple images}.
\newblock In \emph{Proceedings of the AAAI Conference on Artificial Intelligence (AAAI)}, 2023.

\bibitem[Team(2026)]{qwen35blog}
Qwen Team.
\newblock {Qwen3.5: Accelerating Productivity with Native Multimodal Agents}, 2026.

\bibitem[Thrush et~al.(2022)Thrush, Jiang, Bartolo, Singh, Williams, Kiela, and Ross]{thrush2022winoground}
Tristan Thrush, Ryan Jiang, Max Bartolo, Amanpreet Singh, Adina Williams, Douwe Kiela, and Candace Ross.
\newblock {Winoground: Probing vision and language models for visio-linguistic compositionality}.
\newblock In \emph{Proceedings of the Conference on Computer Vision and Pattern Recognition (CVPR)}, 2022.

\bibitem[Tversky(2002)]{tversky2002sketches}
Barbara Tversky.
\newblock {What do sketches say about thinking}.
\newblock In \emph{2002 AAAI Spring Symposium, Sketch Understanding Workshop, Stanford University, AAAI Technical Report SS-02-08}, 2002.

\bibitem[Vo et~al.(2025)Vo, Nguyen, Taesiri, Dang, Nguyen, and Kim]{vo2025visionlanguagemodelsbiased}
An Vo, Khai-Nguyen Nguyen, Mohammad~Reza Taesiri, Vy~Tuong Dang, Anh~Totti Nguyen, and Daeyoung Kim.
\newblock {Vision Language Models are Biased}.
\newblock \emph{arXiv preprint arXiv:2505.23941}, 2025.

\bibitem[Wei et~al.(2022)Wei, Wang, Schuurmans, Bosma, brian ichter, Xia, Chi, Le, and Zhou]{wei2022chain}
Jason Wei, Xuezhi Wang, Dale Schuurmans, Maarten Bosma, brian ichter, Fei Xia, Ed~H. Chi, Quoc~V Le, and Denny Zhou.
\newblock {Chain of Thought Prompting Elicits Reasoning in Large Language Models}.
\newblock In \emph{Advances in Neural Information Processing Systems (NeurIPS)}, 2022.

\bibitem[Wu et~al.(2023)Wu, Zhang, Zhang, Chen, Liao, Wang, Li, Sun, Yan, Zhai, et~al.]{qbench2023}
Haoning Wu, Zicheng Zhang, Erli Zhang, Chaofeng Chen, Liang Liao, Annan Wang, Chunyi Li, Wenxiu Sun, Qiong Yan, Guangtao Zhai, et~al.
\newblock {Q-bench: A benchmark for general-purpose foundation models on low-level vision}.
\newblock \emph{arXiv preprint arXiv:2309.14181}, 2023.

\bibitem[Wu and Xie(2024)]{vstar2023}
Penghao Wu and Saining Xie.
\newblock {V*: Guided visual search as a core mechanism in multimodal llms}.
\newblock In \emph{Proceedings of the Conference on Computer Vision and Pattern Recognition (CVPR)}, 2024.

\bibitem[Wu et~al.(2024)Wu, Chen, Pan, Liu, Liu, Dai, Gao, Ma, Wu, Wang, Xie, Wu, Hu, Wang, Sun, Li, Piao, Guan, Liu, Xie, You, Dong, Yu, Zhang, Zhao, Wang, and Ruan]{wu2024deepseekvl2}
Zhiyu Wu, Xiaokang Chen, Zizheng Pan, Xingchao Liu, Wen Liu, Damai Dai, Huazuo Gao, Yiyang Ma, Chengyue Wu, Bingxuan Wang, Zhenda Xie, Yu Wu, Kai Hu, Jiawei Wang, Yaofeng Sun, Yukun Li, Yishi Piao, Kang Guan, Aixin Liu, Xin Xie, Yuxiang You, Kai Dong, Xingkai Yu, Haowei Zhang, Liang Zhao, Yisong Wang, and Chong Ruan.
\newblock {DeepSeek-VL2: Mixture-of-Experts Vision-Language Models for Advanced Multimodal Understanding}.
\newblock \emph{arXiv preprint arXiv:2412.10302}, 2024.

\bibitem[Xu et~al.(2025)Xu, Shao, Zhang, Gao, Liu, Lei, Meng, Huang, Qiao, and Luo]{xu2025lvlmhum}
Peng Xu, Wenqi Shao, Kaipeng Zhang, Peng Gao, Shuo Liu, Meng Lei, Fanqing Meng, Siyuan Huang, Yu Qiao, and Ping Luo.
\newblock {LVLM-EHub: A Comprehensive Evaluation Benchmark for Large Vision-Language Models}.
\newblock \emph{IEEE Transactions on Pattern Analysis and Machine Intelligence (TPAMI)}, 47\penalty0 (3), 2025.

\bibitem[Yin et~al.(2023)Yin, Wang, Cao, Shi, Liu, Li, Huang, Wang, Sheng, BAI, Shao, and Ouyang]{yin2023lamm}
Zhenfei Yin, Jiong Wang, Jianjian Cao, Zhelun Shi, Dingning Liu, Mukai Li, Xiaoshui Huang, Zhiyong Wang, Lu Sheng, LEI BAI, Jing Shao, and Wanli Ouyang.
\newblock {LAMM: Language-Assisted Multi-Modal Instruction-Tuning Dataset, Framework, and Benchmark}.
\newblock In \emph{Advances in Neural Information Processing Systems (NeurIPS)}, 2023.

\bibitem[Yu et~al.(2024)Yu, Yang, Ren, Li, Wang, Lin, Lin, Liu, Wang, and Wang]{mmvet2024}
Weihao Yu, Zhengyuan Yang, Lingfeng Ren, Linjie Li, Jianfeng Wang, Kevin Lin, Chung-Ching Lin, Zicheng Liu, Lijuan Wang, and Xinchao Wang.
\newblock {Mm-vet v2: A challenging benchmark to evaluate large multimodal models for integrated capabilities}.
\newblock \emph{arXiv preprint arXiv:2408.00765}, 2024.

\bibitem[Yue et~al.(2024)Yue, Ni, Zheng, Zhang, Liu, Zhang, Stevens, Jiang, Ren, Sun, Wei, Yu, Yuan, Sun, Yin, Zheng, Yang, Liu, Huang, Sun, Su, and Chen]{yue2024mmmu}
Xiang Yue, Yuansheng Ni, Tianyu Zheng, Kai Zhang, Ruoqi Liu, Ge Zhang, Samuel Stevens, Dongfu Jiang, Weiming Ren, Yuxuan Sun, Cong Wei, Botao Yu, Ruibin Yuan, Renliang Sun, Ming Yin, Boyuan Zheng, Zhenzhu Yang, Yibo Liu, Wenhao Huang, Huan Sun, Yu Su, and Wenhu Chen.
\newblock {MMMU: A Massive Multi-Discipline Multimodal Understanding and Reasoning Benchmark for Expert AGI}.
\newblock In \emph{Proceedings of the Conference on Computer Vision and Pattern Recognition (CVPR)}, 2024.

\bibitem[Zhang et~al.(2016)Zhang, Goyal, Summers-Stay, Batra, and Parikh]{Zhang_2016_CVPR}
Peng Zhang, Yash Goyal, Douglas Summers-Stay, Dhruv Batra, and Devi Parikh.
\newblock {Yin and Yang: Balancing and Answering Binary Visual Questions}.
\newblock In \emph{Proceedings of the Conference on Computer Vision and Pattern Recognition (CVPR)}, 2016.

\bibitem[Zhang et~al.(2024{\natexlab{a}})Zhang, Dong, Zang, Cao, Qian, Chen, Guo, Duan, Wang, Ouyang, Zhang, Zhang, Li, Gao, Sun, Zhang, Li, Li, Wang, Yan, He, Zhang, Chen, Dai, Qiao, Lin, and Wang]{zhang2024internlmxcomposer25}
Pan Zhang, Xiaoyi Dong, Yuhang Zang, Yuhang Cao, Rui Qian, Lin Chen, Qipeng Guo, Haodong Duan, Bin Wang, Linke Ouyang, Songyang Zhang, Wenwei Zhang, Yining Li, Yang Gao, Peng Sun, Xinyue Zhang, Wei Li, Jingwen Li, Wenhai Wang, Hang Yan, Conghui He, Xingcheng Zhang, Kai Chen, Jifeng Dai, Yu Qiao, Dahua Lin, and Jiaqi Wang.
\newblock {InternLM-XComposer-2.5: A Versatile Large Vision Language Model Supporting Long-Contextual Input and Output}.
\newblock \emph{arXiv preprint arXiv:2407.03320}, 2024{\natexlab{a}}.

\bibitem[Zhang et~al.(2023)Zhang, Han, Liu, Gao, Zhou, Hu, Yan, Lu, Li, and Qiao]{zhang2023llama}
Renrui Zhang, Jiaming Han, Chris Liu, Peng Gao, Aojun Zhou, Xiangfei Hu, Shilin Yan, Pan Lu, Hongsheng Li, and Yu Qiao.
\newblock {Llama-adapter: Efficient fine-tuning of language models with zero-init attention}.
\newblock \emph{arXiv preprint arXiv:2303.16199}, 2023.

\bibitem[Zhang et~al.(2024{\natexlab{b}})Zhang, Unell, Wang, Ghosh, Su, Schmidt, and Yeung-Levy]{zhang2024visually}
Yuhui Zhang, Alyssa Unell, Xiaohan Wang, Dhruba Ghosh, Yuchang Su, Ludwig Schmidt, and Serena Yeung-Levy.
\newblock {Why are visually-grounded language models bad at image classification?}
\newblock \emph{arXiv preprint arXiv:2405.18415}, 2024{\natexlab{b}}.

\bibitem[Zhang et~al.(2025)Zhang, Li, Long, Zhang, Lin, Yang, Xie, Yang, Liu, Lin, Huang, and Zhou]{zhang2025qwen3emb}
Yanzhao Zhang, Mingxin Li, Dingkun Long, Xin Zhang, Huan Lin, Baosong Yang, Pengjun Xie, An Yang, Dayiheng Liu, Junyang Lin, Fei Huang, and Jingren Zhou.
\newblock {Qwen3 Embedding: Advancing Text Embedding and Reranking Through Foundation Models}.
\newblock \emph{arXiv preprint arXiv:2506.05176}, 2025.

\bibitem[Zhu et~al.(2023)Zhu, Chen, Shen, Li, and Elhoseiny]{minigpt4}
Deyao Zhu, Jun Chen, Xiaoqian Shen, Xiang Li, and Mohamed Elhoseiny.
\newblock {MiniGPT-4: Enhancing Vision-Language Understanding with Advanced Large Language Models}.
\newblock In \emph{International Conference on Learning Representations (ICLR)}, 2023.

\bibitem[Zhu et~al.(2025)Zhu, Wang, Chen, Liu, Ye, Gu, Tian, Duan, Su, Shao, Gao, Cui, Wang, Cao, Liu, Wei, Zhang, Wang, Xu, Li, Wang, Deng, Li, He, Jiang, Luo, Wang, He, Shi, Zhang, Shao, He, Xiong, Qu, Sun, Jiao, Lv, Wu, Zhang, Deng, Ge, Chen, Wang, Dou, Lu, Zhu, Lu, Lin, Qiao, Dai, and Wang]{zhu2025internvl3}
Jinguo Zhu, Weiyun Wang, Zhe Chen, Zhaoyang Liu, Shenglong Ye, Lixin Gu, Hao Tian, Yuchen Duan, Weijie Su, Jie Shao, Zhangwei Gao, Erfei Cui, Xuehui Wang, Yue Cao, Yangzhou Liu, Xingguang Wei, Hongjie Zhang, Haomin Wang, Weiye Xu, Hao Li, Jiahao Wang, Nianchen Deng, Songze Li, Yinan He, Tan Jiang, Jiapeng Luo, Yi Wang, Conghui He, Botian Shi, Xingcheng Zhang, Wenqi Shao, Junjun He, Yingtong Xiong, Wenwen Qu, Peng Sun, Penglong Jiao, Han Lv, Lijun Wu, Kaipeng Zhang, Huipeng Deng, Jiaye Ge, Kai Chen, Limin Wang, Min Dou, Lewei Lu, Xizhou Zhu, Tong Lu, Dahua Lin, Yu Qiao, Jifeng Dai, and Wenhai Wang.
\newblock {InternVL3: Exploring Advanced Training and Test-Time Recipes for Open-Source Multimodal Models}.
\newblock \emph{arXiv preprint arXiv:2504.10479}, 2025.

\end{thebibliography}
}

\appendix
\clearpage
\crefalias{section}{appendix}
\renewcommand{\thesection}{\Alph{section}}

\maketitlesupplementary

\renewcommand\contentsname{} %

\begingroup
\let\clearpage\relax
\vspace{-1cm} %

\startcontents[appendix]
\printcontents[appendix]{l}{1}{\section*{Overview}\setcounter{tocdepth}{2}}
\endgroup

\FloatBarrier
\section{Quantifying Image Complexity}\label{subsec:appendix:complexity}
Measuring visual density or complexity remains a non-trivial challenge for standard metrics \cite{forsythe2011predicting,marin2014examining,mahon2023minimumdescriptionlengthclustering,lin2025what}. To obtain complexity measurements of our dataset, we chose to quantify complexity through relative, pairwise comparisons across the full set of 150 images. We automate this process using Qwen3.5-27B (thinking) \cite{qwen35blog} to minimize human bias, sampling responses at temperature $1.0$ ($p=0.95, k=20, penalty=1.0$). The model was provided with the following prompt:
\begin{center}
\small
    \begin{tcolorbox}[colback=gray!5!white,colframe=gray!50!black,left=2pt,right=2pt,top=2pt,bottom=2pt,title={Prompt for Image Complexity (Qwen3.5-27B)}]
    \texttt{<Image 1>}\texttt{<Image 2>} You are an expert in image analysis. Given two images, your task is to determine which image has a higher visual density and complexity. Do not attempt to identify the name of the painting. Respond with 'A' if the first image has higher complexity, 'B' if the second image has higher complexity.
    \end{tcolorbox}
\end{center}
The prompt explicitly instructs the model to ignore the identity of the artwork, encouraging a focus on visual composition rather than semantic recognition. While we observed that the model occasionally attempted identification within its reasoning traces, its final selections remained focused on density. 

\begin{figure}
    \centering

\resizebox{\linewidth}{!}{
    \begin{tabular}{@{}cc@{}}
        \includegraphics[width=0.5\linewidth]{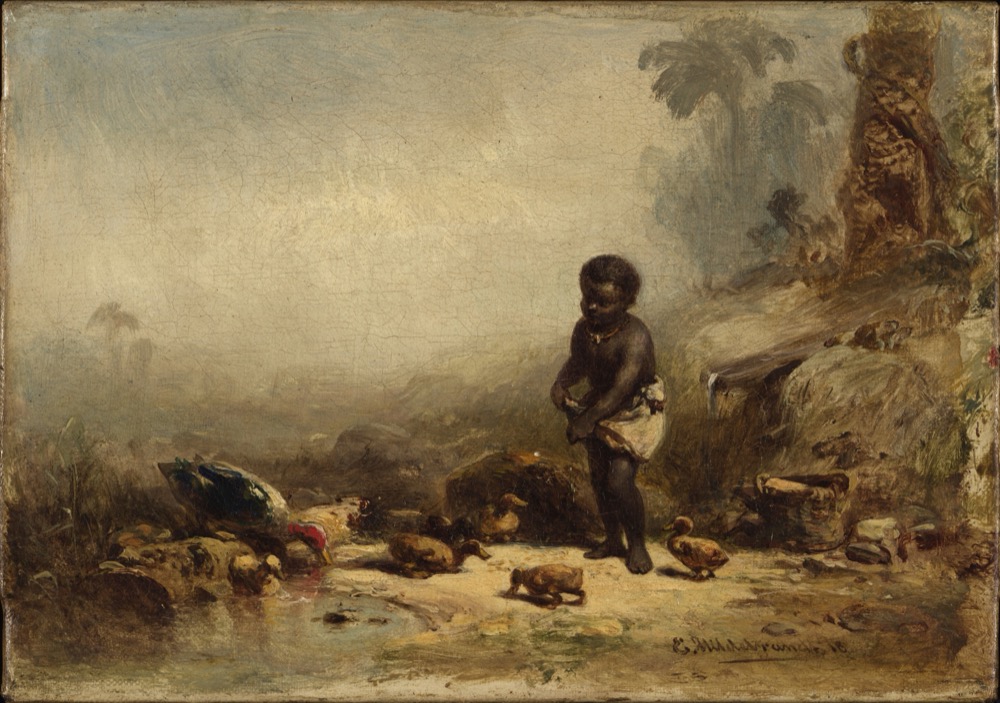} & \includegraphics[width=0.5\linewidth]{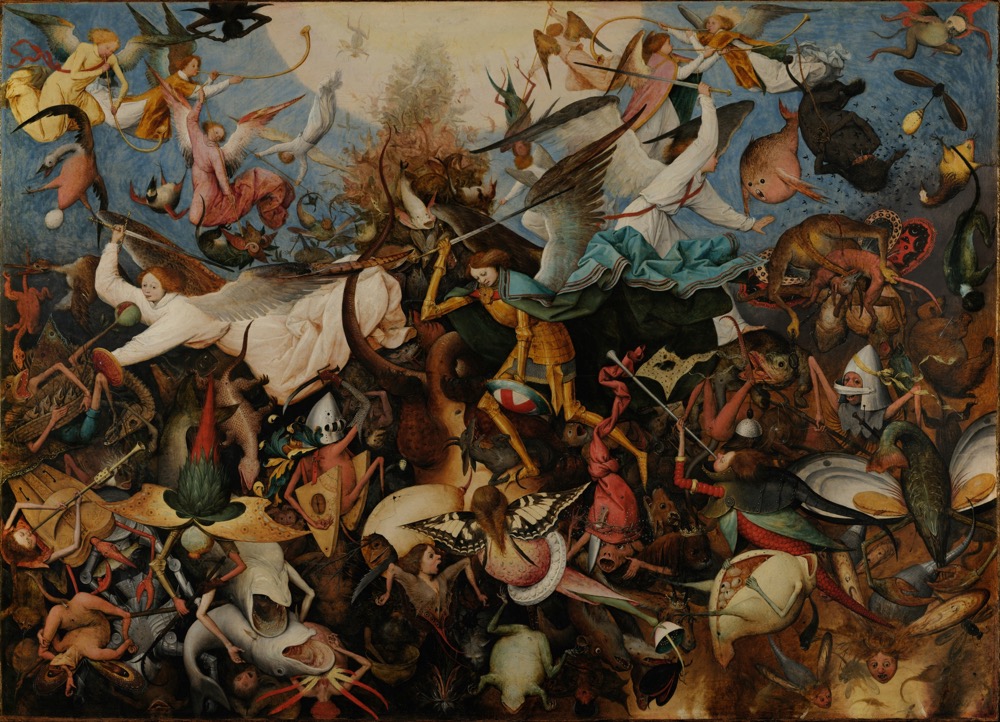}\\
        least complex (0.00 \%) & most complex (99.3 \%)
    \end{tabular}
}

    \caption{\textbf{Extreme ends of complexity distribution.} Shown are the least and most complex images according to our LLM-assisted measurements and their respective win-rate.}
    \label{fig:win_comparison}
\end{figure}

\begin{figure}
    \centering
    \includegraphics[width=\linewidth]{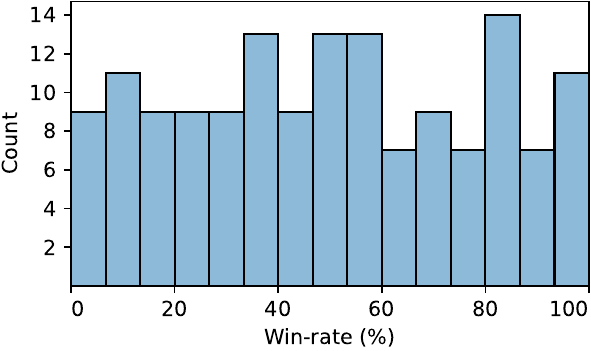}
    \caption{\textbf{Distribution of Image Complexity.} The distribution shows the win-rate over all 150 images in \ourDS.}
    \label{fig:win_rate_dist}
\end{figure}

Ultimately, an image's complexity score is defined as its empirical win-rate: the proportion of pairings in which the model selected it as the more complex image. \cref{fig:win_comparison} illustrates the images at the lowest and highest extremes of this calculated distribution (shown in \cref{fig:win_rate_dist}).

\section{Discussion of Domain Shift}\label{subsec:appendix:domainshift}
\ourDS~consists of paintings, which will display a domain shift compared to natural images. We think of this shift as beneficial because we expect foundation models aiming for human-level perception to also be robust to stylistic abstractions. 

The domain of historical paintings is a good testbed for this kind of robustness, as it offers distinct practical advantages: first, the paintings depict naturalistic scenes and objects before the time of photography. As such, they deviate from digital photos in various ways (style, color, \etc) while still maintaining the \textit{core features} of objects, providing a good trade-off between realism and avoiding the biases of web-crawled data (center-focus, \etc). Second, we assume that this benchmark might be harder to solve than benchmarks consisting of natural images, by just crawling more data (or even generating), as the number of historical paintings is finite and not easily scalable. To improve performance, models will always, to a certain extent, have to be able to generalize. Finally, it is a domain that provides high-resolution images without depth-of-field blur (ensuring details are recognizable across the full scene) and provides a clean copyright status.

\begin{figure}
    \centering
    \includegraphics[width=\linewidth]{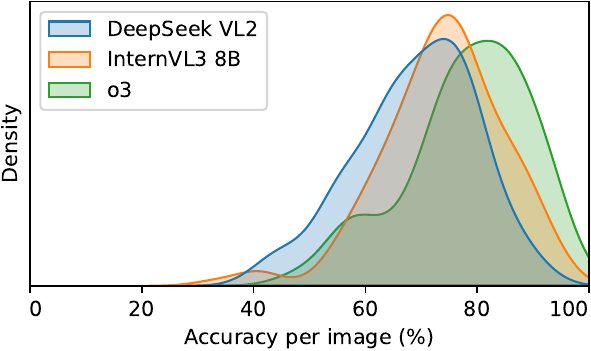}
    \caption{\textbf{Distribution of accuracy per image.} Three diverse models show distributions with clear peaks beyond average accuracy, thus proving that the challenge is caused by the content of the image and question, not the domain.}
    \label{fig:domain_shift}
\end{figure}

Finally, we investigate whether the domain shift may be correlated with the low performance of some models. As shown in the figure (\cref{fig:domain_shift}), the per-image accuracy distributions for 3 distinct models lack the density near zero performance that would characterize a domain shift. Instead, they show a well-above random performance on most images, thus indicating that the performance is not tied to the domain but to the question and visual details. This is further backed by the measurements of {logical consistency} in \cref{fig:logical_consistency}, which show stark differences in performance for tasks differing in the required level of image understanding.

\section{Benchmark Prompts}\label{subsec:appendix:prompts}

We used the following prompts in our main evaluation, depending on the question type (multiple-choice, counting, or OCR):
\begin{center}
\small
    \begin{tcolorbox}[colback=gray!5!white,colframe=gray!50!black,left=2pt,right=2pt,top=2pt,bottom=2pt,title={Default Prompt for Multiple-Choice Questions}]
    \texttt{\{Question\}} Options:\\A. \texttt{\{Option A\}}\\B. \texttt{\{Option B\}}\\$\cdots$\\Answer with the option's letter from the given choices directly.
    \end{tcolorbox}
    
    \begin{tcolorbox}[colback=gray!5!white,colframe=gray!50!black,left=2pt,right=2pt,top=2pt,bottom=2pt,title={Default Prompt for OCR Questions}]
    \texttt{\{Question\}} Answer directly.
    \end{tcolorbox}
    
    \begin{tcolorbox}[colback=gray!5!white,colframe=gray!50!black,left=2pt,right=2pt,top=2pt,bottom=2pt,title={Default Prompt for Counting Questions}]
    \texttt{\{Question\}} Answer with a number directly.
    \end{tcolorbox}
\end{center}

\begin{figure*}
    \centering
    \includegraphics[width=\linewidth]{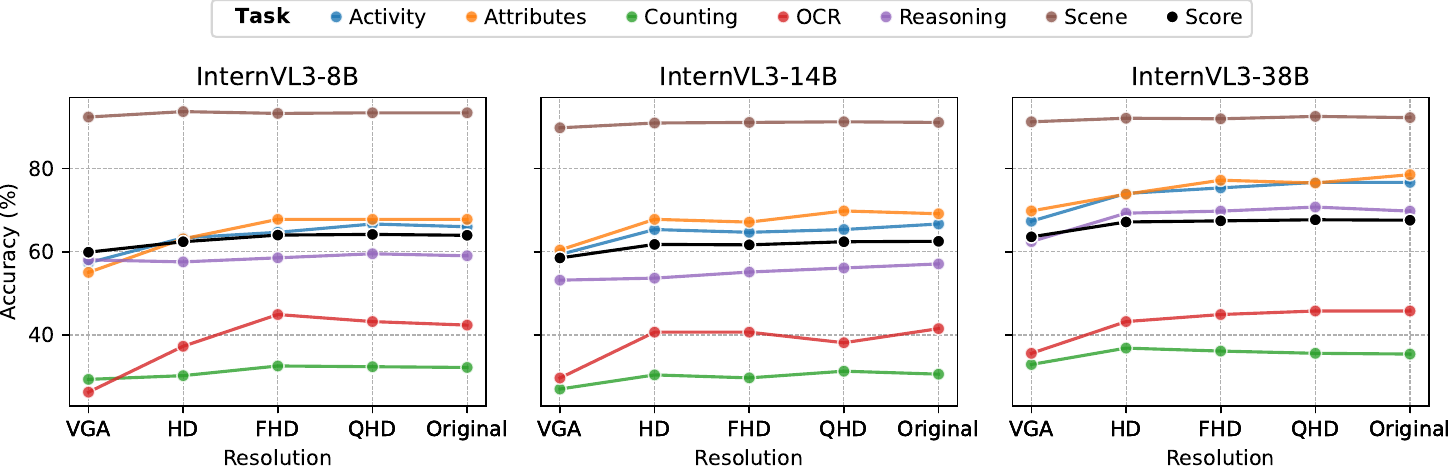}
    \caption{\textbf{Resolution ablation.}}
    \label{fig:resolution_ablation}
\end{figure*}

\section{Ablation of Resolution}\label{appendix:resolution}

We distribute \ourDS~at a resolution that matches the pixels of 4K (with a few outliers). Additionally, we downsampled images to match the number of pixels of VGA ($640 \times 480$ pixels), HD ($1280 \times 720$ pixels), FHD ($1920 \times 1080$ pixels), QHD ($2560 \times 1440$ pixels), and measured task-level performances on various instances of InternVL3 models in comparison to our original resolution. The results are shown in \cref{fig:resolution_ablation}.

Generally, performance improves with resolution, but at a minor rate. However, it is visible that improvements are differently correlated with tasks. Text (especially small one) is poorly compressible, and it is, thus, unsurprising to see a strong correlation between resolution and OCR performance. The opposite is modeled by scene recognition, which, for the most part, is solvable by global features that should be detectable even at extreme compression. This is backed by the lack of significant performance deviation throughout our tested resolutions. For the other tasks, we typically see an increase in performance with resolution, which seems to plateau after Full HD resolution. 

This is likely not a shortcoming of our benchmark, but rather attributed to the model's architecture. By default, InternVL3 splits the input image into at most 12 patches (each $448 \times 448$ pixels) plus a thumbnail \citep{zhu2025internvl3}. Thus, the model only supports a resolution slightly above FHD without downsampling.  While it is possible to increase the number of patches, this significantly increases the inference time and memory. For instance, even for InternVL3-8B, increasing the number of patches from 12 to 40, which should be sufficient to process \ourDS~without downsampling, requires $8 \times 40$ GB GPUs, instead of just one, making such an experiment impossible for us. In theory, we, however, expect model performance to scale with resolution, assuming no downsampling. Consequently, we also expect higher performance using more patches (assuming a sufficient context window and proper training).

\section{Language Bias Detection}\label{appendix:language_bias}

We use Gemini 2.5 Pro with the following prompt to detect language bias:
\begin{center}
\small
    \begin{tcolorbox}[colback=gray!5!white,colframe=gray!50!black,left=2pt,right=2pt,top=2pt,bottom=2pt,title={Prompt for Language Bias Detection (Gemini 2.5 Pro)}]
    \texttt{Below you will find a CSV with an excerpt of questions from a visual question answering benchmark. The benchmark is supposed to be only solvable by looking at the image, however for the questions below, most models are able to guess the correct option (ground\_truth). Your task is to look at each questions, the options, and ground\_truth and to determine if the models were just lucky or there is some kind of shortcut or language bias. Provide an answer and rationale for each question\_id.
    \\\\
    question\_id, question, options, ground\_truth\\
    \{CSV\}
    }
    \end{tcolorbox}
\end{center}
\cref{tab:language_bias} contains the quality control results discussed in \cref{subsec:bench:curation}.

\begin{table*}
\centering
\caption{\textbf{Blind benchmark results.} We benchmark three models on \ourDS~without the images to measure a potential language bias.}
\label{tab:language_bias}
\small
\resizebox{\linewidth}{!}{
\begin{tabular}{@{}lccccccc|ccc|c@{}} 
\toprule
 &                           & \textbf{Activity} & \textbf{Attributes} & \textbf{Counting} & \textbf{OCR} & \textbf{Reasoning} & \textbf{Scene} & \textbf{Easy} & \textbf{Medium} & \textbf{Hard} & \textbf{Total}  \\
\textbf{Model} & \textbf{Params [B]} & (150) & (149) & (559) & (118) & (356) & (1388) & (986) & (1304) & (430) & (2720)\\
\midrule
\textcolor{gray}{\textit{Random Chance}}  & - & \textcolor{gray}{25.0} & \textcolor{gray}{25.0} & \textcolor{gray}{0.0} & \textcolor{gray}{0.0} & \textcolor{gray}{25.0} & \textcolor{gray}{25.0} & \textcolor{gray}{24.5} & \textcolor{gray}{16.7} & \textcolor{gray}{3.7} & \textcolor{gray}{16.0}\\
\textcolor{gray}{\textit{Consistent Chance}} & - & \textcolor{gray}{25.0} & \textcolor{gray}{25.0} & \textcolor{gray}{0.0} & \textcolor{gray}{0.0} & \textcolor{gray}{42.5} & \textcolor{gray}{50.0} & \textcolor{gray}{47.2} & \textcolor{gray}{26.2} & \textcolor{gray}{4.7} & \textcolor{gray}{27.2}\\

\midrule

InternVL3 38B \citep{zhu2025internvl3} & 38 & 30.0 & 34.9 & 15.6 & 0.8 & 36.6 & 24.2 & 32.5 & 24.5 & 8.1 & 22.8 \\
Qwen2.5-VL 32B \citep{bai2025qwen2} & 32 & 32.0 & 26.2 & 8.8 & 0.0 & 29.3 & 38.0 & 39.7 & 25.1 & 6.2 & 24.5 \\
LLaVA-OV 72B \citep{llava-ov} & 72 & 29.3 & 40.3 & 18.1 & 0.8 & 36.1 & 38.6 & 24.6 & 41.1 & 6.5 & 29.2 \\
\bottomrule
\end{tabular}
}
\end{table*}

\section{Performance with Advanced Prompting}

Our evaluation in \cref{sec:experiments} utilizes simple prompts. In this section, we additionally ablate zero-shot chain-of-thought (CoT) \citep{wei2022chain,kojima2022cot} on InternVL3 8B, the strongest 8B model on our benchmark, and an overall strong model. To this end, we modified the prompts as follows:
 \begin{center}
 \small
    \begin{tcolorbox}[colback=gray!5!white,colframe=gray!50!black,left=2pt,right=2pt,top=2pt,bottom=2pt,title={CoT Prompt for Multiple-Choice Questions}]
    \texttt{\{Question\}} Options:\\A. \texttt{\{Option A\}}\\B. \texttt{\{Option B\}}\\$\cdots$\\Think step by step. Answer with the option's letter from the given choices wrapped in <answer></answer>.
    \end{tcolorbox}
    
    \begin{tcolorbox}[colback=gray!5!white,colframe=gray!50!black,left=2pt,right=2pt,top=2pt,bottom=2pt,title={CoT Prompt for OCR Questions}]
    \texttt{\{Question\}} Think step by step. Answer with the extracted text wrapped in <answer></answer>
    \end{tcolorbox}
    
    \begin{tcolorbox}[colback=gray!5!white,colframe=gray!50!black,left=2pt,right=2pt,top=2pt,bottom=2pt,title={CoT Prompt for Counting Questions}]
    \texttt{\{Question\}} Think step by step. Answer with a number wrapped in <answer></answer>
    \end{tcolorbox}
\end{center}
\begin{table*}[]
\centering
\caption{\textbf{Comparison with CoT prompting.}}
\label{table:cot_results}

\resizebox{\linewidth}{!}{
\begin{tabular}{lccccccc|ccc|c} 
\toprule
 & \textbf{Params}                          & \textbf{Activity} & \textbf{Attributes} & \textbf{Counting} & \textbf{OCR} & \textbf{Reasoning} & \textbf{Scene} & \textbf{Easy} & \textbf{Medium} & \textbf{Hard} & \textbf{Total}  \\
\textbf{Model} & \textbf{[B]} & (150) & (149) & (559) & (118) & (356) & (1388) & (986) & (1304) & (430) & (2720)\\
\midrule

InternVL3 38B \citep{zhu2025internvl3} & 38 & 76.7 & 78.5 & 35.4 & 45.8 & 69.8 & 92.2 & 99.7 & 81.8 & 7.2 & 67.6 \\
+ CoT                                  & 38 & \cellcolor{red!25}74.0 & \cellcolor{red!25}69.8 & \cellcolor{red!25}34.5 & \cellcolor{green!25}50.0 & \cellcolor{red!25}62.4 & \cellcolor{red!25}91.4 & \cellcolor{red!25}98.9 & \cellcolor{red!25}77.1 & \cellcolor{green!25}14.4 & \cellcolor{red!25}65.5 \\

\bottomrule
\end{tabular}
}
\end{table*}
The results in \cref{table:cot_results} show that at least for this model, CoT decreased performance on average. However, it significantly improved performance on the hardest split and for OCR. Since CoT prompting is primarily effective in large-scale LLMs \citep{wei2022chain}, we hypothesize that the tested LLM may have been too small to benefit from CoT.

\section{Embedding Space of Benchmark Questions}
\begin{figure}[]
    \centering
    \includegraphics[width=\linewidth]{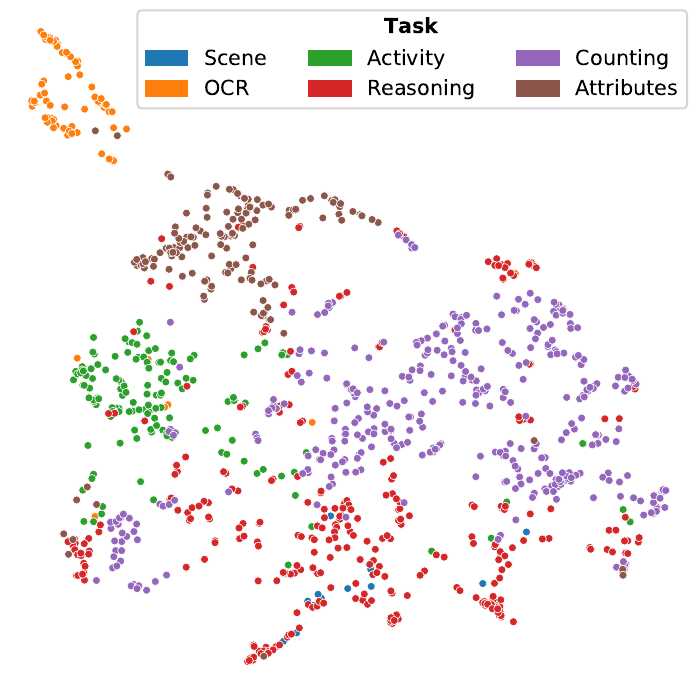}
    \caption{\textbf{Question embeddings.}}
    \label{fig:question_enbeddings}
\end{figure}

We show a UMAP \citep{McInnes2018umap} reduced embedding generated by Qwen3-embedding-4B \citep{zhang2025qwen3emb} of all questions (without answers) colored by task in \cref{fig:question_enbeddings}. A clear separation of tasks is visible, except for the reasoning task, which overlaps with multiple other tasks as intended. The OCR questions form the most disconnected cluster.

\FloatBarrier

\clearpage
\section{Datasheet}

In the following, we provide a datasheet \citep{gebru2021datasheet}. We have anonymized some entries for the review process and will update these upon release.

\newcommand{\dssectionheader}[1]{%
   \noindent\framebox[\columnwidth]{%
      {\textbf{{#1}}}
   }
}

\newcommand{\dsquestion}[1]{%
    {\noindent  {\textbf{#1}}}
}

\newcommand{\dsquestionex}[2]{%
    {\noindent {\textbf{#1}}}
}

\newcommand{\dsanswer}[1]{%
   {\noindent #1 \medskip}
}

\dssectionheader{Motivation}

\dsquestionex{For what purpose was the dataset created?}{Was there a specific task in mind? Was there a specific gap that needed to be filled? Please provide a description.}

\dsanswer{
\ourDS~was created to test basic visual recognition skills of \vlms~in densely populated scenes, as most prior VQA datasets often probe skills of superficial features.
}

\dsquestion{Who created this dataset (e.g., which team, research group) and on behalf of which entity (e.g., company, institution, organization)?}

\dsanswer{
The dataset was created by the authors of this paper on behalf of their institutions.
}

\dsquestionex{Who funded the creation of the dataset?}{If there is an associated grant, please provide the name of the grantor and the grant name and number.}

\dsanswer{
All authors were funded by their respective institutions.
}

\bigskip
\dssectionheader{Composition}

\dsquestionex{What do the instances that comprise the dataset represent (e.g., documents, photos, people, countries)?}{ Are there multiple types of instances (e.g., movies, users, and ratings; people and interactions between them; nodes and edges)? Please provide a description.}

\dsanswer{
The dataset consists of images associated with multiple questions.
}

\dsquestion{How many instances are there in total (of each type, if appropriate)?}

\dsanswer{
The dataset consists of 150 images and a total of 2720 questions.
}

\dsquestionex{Does the dataset contain all possible instances or is it a sample (not necessarily random) of instances from a larger set?}{ If the dataset is a sample, then what is the larger set? Is the sample representative of the larger set (e.g., geographic coverage)? If so, please describe how this representativeness was validated/verified. If it is not representative of the larger set, please describe why not (e.g., to cover a more diverse range of instances, because instances were withheld or unavailable).}

\dsanswer{
The images are a subset of public domain artworks hosted on \url{https://artsandculture.google.com} filtered to display visually complex and dense scenes.
}

\dsquestionex{What data does each instance consist of? “Raw” data (e.g., unprocessed text or images) or features?}{In either case, please provide a description.}

\dsanswer{
Each sample is a collection of the following items:
\begin{itemize}
    \item \texttt{question\_id}: Unique identifier of each question.
    \item \texttt{image}: A PIL JPEG image. Most of our images were resized to match the total pixel count of 4k (3840x2160 px) in different aspect ratios.
    \item \texttt{question}: A question about the image.
    \item \texttt{question\_type}: Type of question. Will be one of choice (response expected to be "A", "B", "C", or "D"), counting (freeform), or ocr (freeform). You can use this information to request a suitable output format.
    \item \texttt{options}: This is the list of options for question\_type=choice and empty otherwise. Please treat the options as answer options A, B, C, D (4 options) or A, B (2 options).
    \item \texttt{difficulty}: Meta-data about the difficulty of the question. One of easy, medium, or hard.
    \item \texttt{category}: Meta-data about the question task. One of activity, attributes, counting, ocr, reasoning, or scene.
    \item \texttt{default\_prompt}: You can use this prompt to stay compliant with our results. It is a simple combination of the question and answers, with some additional output format constraints. This should work well for most models.
\end{itemize}
Further, we provide a database linking each \texttt{image} to its respective \texttt{complexity\_win\_rate}, as defined in \cref{subsec:appendix:complexity}.
}

\dsquestionex{Is there a label or target associated with each instance?}{If so, please provide a description.}

\dsanswer{
Each question is associated with a ground-truth. This ground-truth is hidden from the public to avoid test leakage.
}

\dsquestionex{Is any information missing from individual instances?}{If so, please provide a description, explaining why this information is missing (e.g., because it was unavailable). This does not include intentionally removed information, but might include, e.g., redacted text.}

\dsanswer{
We obfuscate image file names and question IDs to reduce knowledge priors.
}

\dsquestionex{Are relationships between individual instances made explicit (e.g., users’ movie ratings, social network links)?}{If so, please describe how these relationships are made explicit.}

\dsanswer{
The samples in the dataset shall be treated independently.
}

\dsquestionex{Are there recommended data splits (e.g., training, development/validation, testing)?}{If so, please provide a description of these splits, explaining the rationale behind them.}

\dsanswer{
All the samples in our dataset shall be exclusively treated as a test set. We do not provide development sets, as we consider all questions to be solvable with a basic set of skills that should be present in frontier \vlms.
}

\dsquestionex{Are there any errors, sources of noise, or redundancies in the dataset?}{If so, please provide a description.}

\dsanswer{
All questions and ground truths are manually annotated and, thus, may contain errors. To reduce the error rate, we double-checked all questions where multiple models provided wrong answers.
}

\dsquestionex{Is the dataset self-contained, or does it link to or otherwise rely on external resources (e.g., websites, tweets, other datasets)?}{If it links to or relies on external resources, a) are there guarantees that they will exist, and remain constant, over time; b) are there official archival versions of the complete dataset (i.e., including the external resources as they existed at the time the dataset was created); c) are there any restrictions (e.g., licenses, fees) associated with any of the external resources that might apply to a future user? Please provide descriptions of all external resources and any restrictions associated with them, as well as links or other access points, as appropriate.}

\dsanswer{
The dataset is self-contained.
}

\dsquestionex{Does the dataset contain data that might be considered confidential (e.g., data that is protected by legal privilege or by doctor-patient confidentiality, data that includes the content of individuals non-public communications)?}{If so, please provide a description.}

\dsanswer{
No.
}

\dsquestionex{Does the dataset contain data that, if viewed directly, might be offensive, insulting, threatening, or might otherwise cause anxiety?}{If so, please describe why.}

\dsanswer{
The dataset contains samples that show religious beliefs, (partial) nudity, and/or injury and death. 
}

\dsquestionex{Does the dataset relate to people?}{If not, you may skip the remaining questions in this section.}

\dsanswer{
The dataset contains artworks that may depict people.
}

\dsquestionex{Does the dataset identify any subpopulations (e.g., by age, gender)?}{If so, please describe how these subpopulations are identified and provide a description of their respective distributions within the dataset.}

\dsanswer{
The dataset does not identify any subpopulations.
}

\dsquestionex{Is it possible to identify individuals (i.e., one or more natural persons), either directly or indirectly (i.e., in combination with other data) from the dataset?}{If so, please describe how.}

\dsanswer{
Some of the individuals are of historical, biblical, or mythical origin and may be identified. No living individuals can be identified from the dataset.
}

\dsquestionex{Does the dataset contain data that might be considered sensitive in any way (e.g., data that reveals racial or ethnic origins, sexual orientations, religious beliefs, political opinions or union memberships, or locations; financial or health data; biometric or genetic data; forms of government identification, such as social security numbers; criminal history)?}{If so, please provide a description.}

\dsanswer{
No.
}

\bigskip
\dssectionheader{Collection Process}

\dsquestionex{How was the data associated with each instance acquired?}{Was the data directly observable (e.g., raw text, movie ratings), reported by subjects (e.g., survey responses), or indirectly inferred/derived from other data (e.g., part-of-speech tags, model-based guesses for age or language)? If data was reported by subjects or indirectly inferred/derived from other data, was the data validated/verified? If so, please describe how.}

\dsanswer{
Please see \cref{sec:creating}.
}

\dsquestionex{What mechanisms or procedures were used to collect the data (e.g., hardware apparatus or sensor, manual human curation, software program, software API)?}{How were these mechanisms or procedures validated?}

\dsanswer{
Please see \cref{sec:creating}.
}

\dsquestion{If the dataset is a sample from a larger set, what was the sampling strategy (e.g., deterministic, probabilistic with specific sampling probabilities)?}

\dsanswer{
n/a.
}

\dsquestion{Who was involved in the data collection process (e.g., students, crowdworkers, contractors) and how were they compensated (e.g., how much were crowdworkers paid)?}

\dsanswer{
The dataset was collected and annotated by the authors of this paper. No crowdworkers, students, or contractors, \etc, were involved.
}

\dsquestionex{Over what timeframe was the data collected? Does this timeframe match the creation timeframe of the data associated with the instances (e.g., recent crawl of old news articles)?}{If not, please describe the timeframe in which the data associated with the instances was created.}

\dsanswer{
The images were collected between April and May 2025, and annotated and cleaned between May and August 2025.
}

\dsquestionex{Were any ethical review processes conducted (e.g., by an institutional review board)?}{If so, please provide a description of these review processes, including the outcomes, as well as a link or other access point to any supporting documentation.}

\dsanswer{
No.
}

\dsquestionex{Does the dataset relate to people?}{If not, you may skip the remaining questions in this section.}

\dsanswer{
The dataset contains artworks that may depict people.
}

\dsquestion{Did you collect the data from the individuals in question directly, or obtain it via third parties or other sources (e.g., websites)?}

\dsanswer{
n/a.
}

\dsquestionex{Were the individuals in question notified about the data collection?}{If so, please describe (or show with screenshots or other information) how notice was provided, and provide a link or other access point to, or otherwise reproduce, the exact language of the notification itself.}

\dsanswer{
All depicted individuals are no longer alive.
}

\dsquestionex{Did the individuals in question consent to the collection and use of their data?}{If so, please describe (or show with screenshots or other information) how consent was requested and provided, and provide a link or other access point to, or otherwise reproduce, the exact language to which the individuals consented.}

\dsanswer{
n/a.
}

\dsquestionex{If consent was obtained, were the consenting individuals provided with a mechanism to revoke their consent in the future or for certain uses?}{If so, please provide a description, as well as a link or other access point to the mechanism (if appropriate).}

\dsanswer{
n/a.
}

\dsquestionex{Has an analysis of the potential impact of the dataset and its use on data subjects (e.g., a data protection impact analysis) been conducted?}{If so, please provide a description of this analysis, including the outcomes, as well as a link or other access point to any supporting documentation.}

\dsanswer{
n/a.
}

\bigskip
\dssectionheader{Preprocessing/cleaning/labeling}

\dsquestionex{Was any preprocessing/cleaning/labeling of the data done (e.g., discretization or bucketing, tokenization, part-of-speech tagging, SIFT feature extraction, removal of instances, processing of missing values)?}{If so, please provide a description. If not, you may skip the remainder of the questions in this section.}

\dsanswer{
Yes, see \cref{sec:creating}.
}

\dsquestionex{Was the “raw” data saved in addition to the preprocessed/cleaned/labeled data (e.g., to support unanticipated future uses)?}{If so, please provide a link or other access point to the “raw” data.}

\dsanswer{
The raw data can be requested from the authors.
}

\dsquestionex{Is the software used to preprocess/clean/label the instances available?}{If so, please provide a link or other access point.}

\dsanswer{
The images were obtained using \url{https://github.com/lovasoa/dezoomify-rs}. All further processing scripts were developed by the authors and are not available publicly.
}

\bigskip
\dssectionheader{Uses}

\dsquestionex{Has the dataset been used for any tasks already?}{If so, please provide a description.}

\dsanswer{
The dataset has been used to evaluate basic visual skills of frontier \vlms~in \cref{sec:experiments}.
}

\dsquestionex{Is there a repository that links to any or all papers or systems that use the dataset?}{If so, please provide a link or other access point.}

\dsanswer{
We will list relevant papers at \url{https://github.com/paulgavrikov/visualoverload}. We encourage authors to contact us to list their works.
} %

\dsquestion{What (other) tasks could the dataset be used for?}

\dsanswer{
The dataset is primarily designed for visual question answering (VQA), but we encourage users to apply it to other tasks as desired.
}

\dsquestionex{Is there anything about the composition of the dataset or the way it was collected and preprocessed/cleaned/labeled that might impact future uses?}{For example, is there anything that a future user might need to know to avoid uses that could result in unfair treatment of individuals or groups (e.g., stereotyping, quality of service issues) or other undesirable harms (e.g., financial harms, legal risks) If so, please provide a description. Is there anything a future user could do to mitigate these undesirable harms?}

\dsanswer{
No.
}

\dsquestionex{Are there tasks for which the dataset should not be used?}{If so, please provide a description.}

\dsanswer{
This dataset is released exclusively for academic research and educational use. It must not be applied to purposes that could lead to harm, including surveillance, discrimination, exploitation, harassment, or the generation of misleading or offensive content. Users are expected to uphold the highest standards of research integrity and ethics, and to ensure that their work with this dataset aligns with responsible AI principles.
}

\bigskip
\dssectionheader{Distribution}

\dsquestionex{Will the dataset be distributed to third parties outside of the entity (e.g., company, institution, organization) on behalf of which the dataset was created?}{If so, please provide a description.}

\dsanswer{
The dataset is publicly available.
}

\dsquestionex{How will the dataset be distributed (e.g., tarball on website, API, GitHub)}{Does the dataset have a digital object identifier (DOI)?}

\dsanswer{
The dataset is distributed through HuggingFace datasets, which currently uses a PyArrow format.
}

\dsquestion{When will the dataset be distributed?}

\dsanswer{
The dataset is immediately distributed through: \url{https://huggingface.co/datasets/paulgavrikov/visualoverload}.
} %

\dsquestionex{Will the dataset be distributed under a copyright or other intellectual property (IP) license, and/or under applicable terms of use (ToU)?}{If so, please describe this license and/or ToU, and provide a link or other access point to, or otherwise reproduce, any relevant licensing terms or ToU, as well as any fees associated with these restrictions.}

\dsanswer{
The dataset is distributed under the Creative Commons Attribution-ShareAlike 4.0 International license without any further terms of use.
}

\dsquestionex{Have any third parties imposed IP-based or other restrictions on the data associated with the instances?}{If so, please describe these restrictions, and provide a link or other access point to, or otherwise reproduce, any relevant licensing terms, as well as any fees associated with these restrictions.}

\dsanswer{
No.
}

\dsquestionex{Do any export controls or other regulatory restrictions apply to the dataset or to individual instances?}{If so, please describe these restrictions, and provide a link or other access point to, or otherwise reproduce, any supporting documentation.}

\dsanswer{
No.
}

\bigskip
\dssectionheader{Maintenance}

\dsquestion{Who will be supporting/hosting/maintaining the dataset?}

\dsanswer{
The authors will be supporting/hosting/maintaining the dataset.
}

\dsquestion{How can the owner/curator/manager of the dataset be contacted (e.g., email address)?}

\dsanswer{
The authors can be contacted via GitHub issues at: \url{https://github.com/paulgavrikov/visualoverload/issues}. %
}

\dsquestionex{Is there an erratum?}{If so, please provide a link or other access point.}

\dsanswer{
No.
}

\dsquestionex{Will the dataset be updated (e.g., to correct labeling errors, add new instances, delete instances)?}{If so, please describe how often, by whom, and how updates will be communicated to users (e.g., mailing list, GitHub)?}

\dsanswer{
The dataset will not be modified to ensure comparability of results. Corrected or derived datasets will be released independently.
}

\dsquestionex{If the dataset relates to people, are there applicable limits on the retention of the data associated with the instances (e.g., were individuals in question told that their data would be retained for a fixed period of time and then deleted)?}{If so, please describe these limits and explain how they will be enforced.}

\dsanswer{
n/a.
}

\dsquestionex{Will older versions of the dataset continue to be supported/hosted/maintained?}{If so, please describe how. If not, please describe how its obsolescence will be communicated to users.}

\dsanswer{
The dataset will remain available as long as it continues to be hosted by the third-party platforms on which it is stored.}

\dsquestionex{If others want to extend/augment/build on/contribute to the dataset, is there a mechanism for them to do so?}{If so, please provide a description. Will these contributions be validated/verified? If so, please describe how. If not, why not? Is there a process for communicating/distributing these contributions to other users? If so, please provide a description.}

\dsanswer{
Users can extend/augment/build upon the dataset, but must publish their new work as a standalone derivative. We kindly request that users communicate any releases to the authors.
}

\onecolumn
\section{Image References}

\begin{longtable}{@{}p{4cm}p{8cm}p{2.5cm}p{1cm}@{}}
\caption{\textbf{List of artworks.} This table contains all artworks present in \ourDS~in random order. The metadata is taken from the references with minor postprocessing by us.}\label{tab:artworks}\\

\toprule
\textbf{Creator} & \textbf{Title} & \textbf{Date} & \textbf{URL}\\
\midrule

Unknown & Wood of the Philosophers & 1800/1830 & \href{https://artsandculture.google.com/asset/wood-of-the-philosophers-unknown/\_QGC8dr2sfW4bQ}{Link} \\\midrule
Pieter Aertsen & The Fat Kitchen. An Allegory & 1565-1575 & \href{https://artsandculture.google.com/asset/the-fat-kitchen-an-allegory-pieter-aertsen/MwGNbZwwObkz6Q}{Link} \\\midrule
Unknown & A handscroll painting of the porcelain production process (right half) &  early 19th century & \href{https://artsandculture.google.com/asset/a-handscroll-painting-of-the-porcelain-production-process-early-19th-century-right-half/gwEBFS\_KkROLLA}{Link} \\\midrule
Avercamp, Hendrick & Enjoying the Ice & ca.~1615-1620 & \href{https://artsandculture.google.com/asset/enjoying-the-ice-avercamp-hendrick/KAFNo7lj8DQnCw}{Link} \\\midrule
Charles M. Russell & The Medicine Man & 1908 & \href{https://artsandculture.google.com/asset/the-medicine-man-charles-m-russell/wQHH2IMngImn\_g}{Link} \\\midrule
Pieter van der Heyden after Pieter Bruegel the Elder  & The Big Fish Eat the Little Fish & published 1557 & \href{https://artsandculture.google.com/asset/the-big-fish-eat-the-little-fish-pieter-van-der-heyden-after-pieter-bruegel-the-elder-after-hieronymus-bosch/XQGeTHeLiB63fw}{Link} \\\midrule
Charles Fairfax Murray  & Allegory of Good Government, after Ambrogio Lorenzetti & 1873 & \href{https://artsandculture.google.com/asset/allegory-of-good-government-after-ambrogio-lorenzetti-charles-fairfax-murray-1849-1919/1QGsCmxWjYYUxg}{Link} \\\midrule
Philip Galle after Pieter Bruegel the Elder & Prudence & published 1559 & \href{https://artsandculture.google.com/asset/prudence-philip-galle-after-pieter-bruegel-the-elder/8gHY\_xYOTq0w2Q}{Link} \\\midrule
Pieter Bruegel the Elder, Frans Hogenberg & The Kermis at Hoboken & ca.~1559 & \href{https://artsandculture.google.com/asset/the-kermis-at-hoboken-pieter-bruegel-the-elder-frans-hogenberg/0QHaGqRzKbh2Ug}{Link} \\\midrule
Joan Antigó, Honorat Borrassà i Francesc Vergós & Altarpiece of Saint Miquel de Castelló d'Empúries (detail) & 1448 & \href{https://artsandculture.google.com/asset/altarpiece-of-saint-miquel-de-castell\%C3\%B3-d-emp\%C3\%BAries-detail-joan-antig\%C3\%B3-honorat-borrass\%C3\%A0-i-francesc-verg\%C3\%B3s/ZQF2oGY7n-2r6A}{Link} \\\midrule
Greek artist from the end of the 18th century & St. George & 1798/1798 & \href{https://artsandculture.google.com/asset/st-george-greek-artist-from-the-end-of-the-18th-century/nwErY5KB9ggFhg}{Link} \\\midrule
Pieter Bruegel the Elder, Pieter van der Heyden, Hieronymus Cock & Anger (Ira) from The Seven Deadly Sins & 1558 & \href{https://artsandculture.google.com/asset/anger-ira-from-the-seven-deadly-sins-pieter-bruegel-the-elder-pieter-van-der-heyden-hieronymus-cock/2AFmp3WSNAR1WQ}{Link} \\\midrule
Dirck Franchoisz Hals,  Dirck van Delen & Festive Company in a Renaissance Room & 1628 & \href{https://artsandculture.google.com/asset/festive-company-in-a-renaissance-room-dirck-franchoisz-hals-and-dirck-van-delen/-wHflPHndLOcfQ}{Link} \\\midrule
Philips Galle, Pieter Bruegel the Elder, Hieronymus Cock & Charity (Charitas) from The Virtues & 1559 & \href{https://artsandculture.google.com/asset/charity-charitas-from-the-virtues-philips-galle-pieter-bruegel-the-elder-hieronymus-cock/OAEfKdNpVRlr\_Q}{Link} \\\midrule
Unknown & A Sunday on La Grande Jatte & 1884-1886 & \href{https://artsandculture.google.com/asset/a-sunday-on-la-grande-jatte/twGyqq52R-lYpA}{Link} \\\midrule
Charles William Sharpe & The Death of Nelson at the Battle of Trafalgar & 1806/1876 & \href{https://artsandculture.google.com/asset/the-death-of-nelson-at-the-battle-of-trafalgar-charles-william-sharpe/jgEXGryPmA2Zmw}{Link} \\\midrule
Ast, Balthasar van der & Still Life with Fruit and Flowers & 1620-1621 & \href{https://artsandculture.google.com/asset/still-life-with-fruit-and-flowers-ast-balthasar-van-der/eAFqfotCnikxRQ}{Link} \\\midrule
School of Canaletto & St. Marks, Venice & unknown & \href{https://artsandculture.google.com/asset/st-marks-venice-school-of-canaletto/XgFbjtjHCEmXDA}{Link} \\\midrule
Pieter Bruegel the Elder & The Sermon of Saint John the Baptist & 1566 & \href{https://artsandculture.google.com/asset/the-sermon-of-saint-john-the-baptist-pieter-bruegel-the-elder/eQFtxANEO2mvnQ}{Link} \\\midrule
Pieter Aertsen & Market Scene & 1569 & \href{https://artsandculture.google.com/asset/market-scene-0000/vwE63xkF3J9A2g}{Link} \\\midrule
Thomas Matthews Rooke  & Washing Sheds at Chartres & 1885 & \href{https://artsandculture.google.com/asset/washing-sheds-at-chartres-thomas-matthews-rooke-1842-1942/FgHnV8ccmjHmhw}{Link} \\\midrule
Severin Roesen & Still Life of Flowers and Fruit with a River Landscape in the Distance & 1867 & \href{https://artsandculture.google.com/asset/still-life-of-flowers-and-fruit-with-a-river-landscape-in-the-distance-severin-roesen/xgEBpcSCAkx-gA}{Link} \\\midrule
Pere Mates & Final Judgment. Altarpiece of Santa Maria de Segueró (Garrotxa) & 1500/1550 & \href{https://artsandculture.google.com/asset/final-judgment-altarpiece-of-santa-maria-de-seguer\%C3\%B3-garrotxa-pere-mates/3AGkYyW7Jpnz8g}{Link} \\\midrule
Steen, Jan Havicksz & Villagers Merrymaking Outside an Inn & 1652 & \href{https://artsandculture.google.com/asset/villagers-merrymaking-outside-an-inn-steen-jan-havicksz/awFQDHdiTftMeA}{Link} \\\midrule
Hieronymus Bosch & Ecce Homo & 1500 & \href{https://artsandculture.google.com/asset/ecce-homo-hieronymus-bosch/DQFFBFYxlip-xw}{Link} \\\midrule
Jan Steen & Beware of Luxury (“In Weelde Siet Toe”) & 1663 & \href{https://artsandculture.google.com/asset/beware-of-luxury-\%E2\%80\%9Cin-weelde-siet-toe\%E2\%80\%9D-jan-steen/iAEDJelKemoXnA}{Link} \\\midrule
Unknown & Christ in the House of Martha and Mary & 1553 & \href{https://artsandculture.google.com/asset/christ-in-the-house-of-martha-and-mary/0QFDb6g5E0YIbg}{Link} \\\midrule
Ostade, Adriaen van & Peasants in an Interior & 1661 & \href{https://artsandculture.google.com/asset/peasants-in-an-interior-ostade-adriaen-van/2AE9O5NxUCRjrw}{Link} \\\midrule
Pieter Bruegel the Elder & Desidia (Sloth) & 1557 & \href{https://artsandculture.google.com/asset/desidia-sloth-1557-pieter-bruegel-the-elder/UAGiJPlAtmzqqw}{Link} \\\midrule
Avercamp, Hendrick & Enjoying the Ice near a Town & ca.~1620 & \href{https://artsandculture.google.com/asset/enjoying-the-ice-near-a-town-avercamp-hendrick/kQHg9-HRDLAW1Q}{Link} \\\midrule
Rijn, Rembrandt van & The Night Watch & 1642 & \href{https://artsandculture.google.com/asset/the-night-watch-rijn-rembrandt-van/eQEojRwTdypUKA}{Link} \\\midrule
Ditlev Blunck & Danish artists at the Osteria La Gensola in Rome & 1837 & \href{https://artsandculture.google.com/asset/danish-artists-at-the-osteria-la-gensola-in-rome-ditlev-blunck/UwFsvTTIx6ff8Q}{Link} \\\midrule
Konstantin Makovsky & A Boyar Wedding Feast & 1883 & \href{https://artsandculture.google.com/asset/a-boyar-wedding-feast-konstantin-makovsky/igHPP\_wsY9gKZQ}{Link} \\\midrule
Unknown & The trial of the Neptune's seamen & 1807 & \href{https://artsandculture.google.com/asset/the-trial-of-the-neptune-s-seamen/zAGeMC\_4fXiPxw}{Link} \\\midrule
Albrecht Altdorfer & Christ taking Leave of his Mother & probably 1520 & \href{https://artsandculture.google.com/asset/christ-taking-leave-of-his-mother-albrecht-altdorfer/GwGLtHq4IJbiEQ}{Link} \\\midrule
Baines, Thomas & Kaffirs and Rebel Hottentotts Attacking a Wagon Train & 1851/1852 & \href{https://artsandculture.google.com/asset/kaffirs-and-rebel-hottentotts-attacking-a-wagon-train-baines-thomas/9AHMI5NxcPMx1A}{Link} \\\midrule
Unknown & Bucentaur's return to the pier by the Palazzo Ducale & 1728/1729 & \href{https://artsandculture.google.com/asset/bucentaur-s-return-to-the-pier-by-the-palazzo-ducale/mwEV7sO9uSFCpw}{Link} \\\midrule
Neer, Aert van der & Winter Landscape near a Town with Kolf Players and Horse-Drawn Sleighs & ca.~1650-1655 & \href{https://artsandculture.google.com/asset/winter-landscape-near-a-town-with-kolf-players-and-horse-drawn-sleighs-neer-aert-van-der/4QEjKBdiTcCT\_Q}{Link} \\\midrule
Philip Galle after Pieter Bruegel the Elder & Faith & published 1559 & \href{https://artsandculture.google.com/asset/faith-philip-galle-after-pieter-bruegel-the-elder/UAHf4ymagaTkCA}{Link} \\\midrule
Francesco Hayez & Pope Urban II Preaching the First Crusade in the Square of Clermont & 1835/1835 & \href{https://artsandculture.google.com/asset/pope-urban-ii-preaching-the-first-crusade-in-the-square-of-clermont-francesco-hayez/XwE0R12q1msw6A}{Link} \\\midrule
Pieter Bruegel the Elder & The Fall of the Rebel Angels & 1562 & \href{https://artsandculture.google.com/asset/the-fall-of-the-rebel-angels-pieter-bruegel-the-elder/ewEs\_8lOXkz7tQ}{Link} \\\midrule
Atelier de Paris & Psyché rapporte la laine des brebis & 1650 & \href{https://artsandculture.google.com/asset/psych\%C3\%A9-rapporte-la-laine-des-brebis-atelier-de-paris/jAGnBn5TZCVF7Q}{Link} \\\midrule
Hals, Dirck & The Fête champêtre & 1627 & \href{https://artsandculture.google.com/asset/the-f\%C3\%AAte-champ\%C3\%AAtre-hals-dirck/uAHjwWyuG1QV4A}{Link} \\\midrule
Baines, Henry & Fisherfleet Looking East & 1823/1894 & \href{https://artsandculture.google.com/asset/fisherfleet-looking-east-baines-henry/UQEhuQHMXHjasg}{Link} \\\midrule
Jan Cornelisz. Vermeyen & The Spanish Brothel & 1545 & \href{https://artsandculture.google.com/asset/the-spanish-brothel-jan-cornelisz-vermeyen/JgHRsYoggbGlgA}{Link} \\\midrule
Marià Vayreda i Vila & Gambeto dance in Riudaura & 1890 & \href{https://artsandculture.google.com/asset/gambeto-dance-in-riudaura-mari\%C3\%A0-vayreda-i-vila/PQG16gWMZHvdmw}{Link} \\\midrule
Paolo De Matteis & St. Nicolas of Bari Felling a Tree Inhabited by Demons & 1727/1727 & \href{https://artsandculture.google.com/asset/st-nicolas-of-bari-felling-a-tree-inhabited-by-demons-paolo-de-matteis/0gGt8CMxkhj83A}{Link} \\\midrule
Philip Galle,  Pieter Bruegel  & The Resurrection of Christ & ca.~1562 & \href{https://artsandculture.google.com/asset/the-resurrection-of-christ-philip-galle-flemish-1537-1612/ZAFtuXKvElZlvQ}{Link} \\\midrule
Pieter Aertsen & Market Scene & 1550 & \href{https://artsandculture.google.com/asset/market-scene-0000/HgF9oWa8DRbAvg}{Link} \\\midrule
Pieter Aertsen & Christ with Mary and Martha & 1552 & \href{https://artsandculture.google.com/asset/christ-with-mary-and-martha-0000/fQHResTtbF-Cpw}{Link} \\\midrule
Philip Galle after Pieter Bruegel the Elder & The Parable of the Wise and Foolish Virgins & ca.~1560/1563 & \href{https://artsandculture.google.com/asset/the-parable-of-the-wise-and-foolish-virgins-philip-galle-after-pieter-bruegel-the-elder/\_gEDoZzUrMXVtQ}{Link} \\\midrule
Unknown & Dragon Boats at Aberdeen Hong Kong showing Careening Island & 1923 & \href{https://artsandculture.google.com/asset/dragon-boats-at-aberdeen-hong-kong-showing-careening-island/mQFrnzFdQTMzUQ}{Link} \\\midrule
Pieter Bruegel the Elder, Pieter van der Heyden, Hieronymus Cock & Avarice (Avaritia), from the series The Seven Deadly Sins & 1558 & \href{https://artsandculture.google.com/asset/avarice-avaritia-from-the-series-the-seven-deadly-sins-pieter-bruegel-the-elder-pieter-van-der-heyden-hieronymus-cock/hAHQYpGuwFfn6g}{Link} \\\midrule
Hendrick Avercamp & A Scene on the Ice & ca.~1625 & \href{https://artsandculture.google.com/asset/a-scene-on-the-ice-hendrick-avercamp/0AEcjdvrkjPfdg}{Link} \\\midrule
Attributed to Jan van Belcamp & The Great Picture & 1646 & \href{https://artsandculture.google.com/asset/the-great-picture-attributed-to-jan-van-belcamp/ugHL4\_ozVj1f3g}{Link} \\\midrule
Aleksander Ivanov & The Apparition of Christ to the People (The Apparition of the Messiah) & 1837-1857 & \href{https://artsandculture.google.com/asset/the-apparition-of-christ-to-the-people-the-apparition-of-the-messiah-aleksander-ivanov/lgGqUffODe21kA}{Link} \\\midrule
Hendrick Avercamp & Frozen River with Skaters & 1620s & \href{https://artsandculture.google.com/asset/frozen-river-with-skaters-hendrick-avercamp/LQG-vyH-NxHYmA}{Link} \\\midrule
Jan Rost & The Pharaoh Welcomes Joseph & 1553 & \href{https://artsandculture.google.com/asset/the-pharaoh-welcomes-joseph-jan-rost-tapestry-factory-on-a-bronzino-cartoon/QQGDekzrHzTexw}{Link} \\\midrule
Aertsen, Pieter & Wing of an Altarpiece with Adoration of the Magi, on the reverse is Presentation in the Temple & 1560-1565 & \href{https://artsandculture.google.com/asset/wing-of-an-altarpiece-with-adoration-of-the-magi-on-the-reverse-is-presentation-in-the-temple-0001/iQHUCNGPbmNBRQ}{Link} \\\midrule
Van Aachen, Hans & The Rape of Proserpine & 1589 & \href{https://artsandculture.google.com/asset/the-rape-of-proserpine-van-aachen-hans/mAFy0EASgZpp7Q}{Link} \\\midrule
Avercamp, Hendrick & Winter Landscape with Ice Skaters & ca.~1608 & \href{https://artsandculture.google.com/asset/winter-landscape-with-ice-skaters-avercamp-hendrick/5AE9Ec9G0xa7bg}{Link} \\\midrule
Estevão Silva & Untitled & 1887/1887 & \href{https://artsandculture.google.com/asset/untitled-estev\%C3\%A3o-silva/EAGNFYpEP900ww}{Link} \\\midrule
Pieter Bruegel,  J. Liefrinck,  H. Hondius  & The fat kitchen & 1563 & \href{https://artsandculture.google.com/asset/the-fat-kitchen-pieter-bruegel-designer-j-liefrinck-engraver-and-h-hondius-publisher/AAHk-UZEbKeCDg}{Link} \\\midrule
Unknown & Moses descends from Mount Siniai with the Ten Commandments & 1662 & \href{https://artsandculture.google.com/asset/moses-descends-from-mount-siniai-with-the-ten-commandments/xAFZ7Aywwmfq-g}{Link} \\\midrule
Unknown & British forces receiving Commissioner Keying at Canton's British Factories for conference with Sir J. F. Davies & 1847 & \href{https://artsandculture.google.com/asset/british-forces-receiving-commissioner-keying-at-canton-s-british-factories-for-conference-with-sir-j-f-davies/fQEi7VeEoftI6w}{Link} \\\midrule
Unknown & Le peuple rend les honneurs à Psyché & 1650 & \href{https://artsandculture.google.com/asset/le-peuple-rend-les-honneurs-\%C3\%A0-psych\%C3\%A9/LQHvDW0vrWRNTA}{Link} \\\midrule
Pieter Bruegel the Elder & Peasant Wedding & 1566-1569 & \href{https://artsandculture.google.com/asset/peasant-wedding-pieter-bruegel-the-elder/hgGvote2WI8P3w}{Link} \\\midrule
Jan Miense Molenaer & Self-Portrait with Family Members & 1630/1640 & \href{https://artsandculture.google.com/asset/self-portrait-with-family-members-jan-miense-molenaer/LAG\_zAV0faspzA}{Link} \\\midrule
Philip Galle after Pieter Bruegel the Elder & Temperance & published 1559 & \href{https://artsandculture.google.com/asset/temperance-philip-galle-after-pieter-bruegel-the-elder/mgHbOKjfWQxo9Q}{Link} \\\midrule
Anonymous & Christ as the Good Shepherd & 1505 & \href{https://artsandculture.google.com/asset/christ-as-the-good-shepherd-anonymous/cwGEn6oNmW4ZMQ}{Link} \\\midrule
Dirck Jacobsz Vellert & The Flood & 1544 & \href{https://artsandculture.google.com/asset/the-flood-dirck-jacobsz-vellert/VQG9NXHscRYK8Q}{Link} \\\midrule
Cornelis Cornelisz van Haarlem & The Golden Age (Bacchanal) or the Garden of Love & 1614 & \href{https://artsandculture.google.com/asset/the-golden-age-bacchanal-or-the-garden-of-love-cornelis-cornelisz-van-haarlem/1QGYgrDvPFelNQ}{Link} \\\midrule
Aertsen, Pieter & The Adoration of the Magi & ca.~1560 & \href{https://artsandculture.google.com/asset/the-adoration-of-the-magi-0001/LAHmJwzwFE-DPA}{Link} \\\midrule
William Duffield & Still Life & 1859 & \href{https://artsandculture.google.com/asset/still-life-william-duffield/FAHKKkJHeMlcAg}{Link} \\\midrule
Balthasar van der Ast & Still Life of Flowers, Fruit, Shells, and Insects & About 1629 & \href{https://artsandculture.google.com/asset/still-life-of-flowers-fruit-shells-and-insects-balthasar-van-der-ast/nwGYqkPIspmBow}{Link} \\\midrule
Albrecht Dürer & Feast of Rose Garlands & 1506 & \href{https://artsandculture.google.com/asset/feast-of-rose-garlands-albrecht-d\%C3\%BCrer/ngGI\_fW-CrcSrw}{Link} \\\midrule
Jan Steen & The Dancing Couple & 1663 & \href{https://artsandculture.google.com/asset/the-dancing-couple-jan-steen/MQE2fyQk7CYX-A}{Link} \\\midrule
Jan Steen & The Worship of the Golden Calf & ca.~1672-1675 & \href{https://artsandculture.google.com/asset/the-worship-of-the-golden-calf-jan-steen/PwFItCgdISPGwQ}{Link} \\\midrule
Carl Bloch & In a Roman Osteria & 1866 & \href{https://artsandculture.google.com/asset/in-a-roman-osteria-carl-bloch/QAFizpSvU6qx6g}{Link} \\\midrule
Avercamp, Hendrick & Ice-Skating in a Village & ca.~1610 & \href{https://artsandculture.google.com/asset/ice-skating-in-a-village-avercamp-hendrick/7QF6m8xRix7ZGQ}{Link} \\\midrule
After Pieter Bruegel the Elder  & The Festival of Fools & after 1570 & \href{https://artsandculture.google.com/asset/the-festival-of-fools-after-pieter-bruegel-the-elder-netherlandish-breda-ca-1525\%E2\%80\%931569-brussels/YgGYAj9wOPXASA}{Link} \\\midrule
Pieter van der Heyden after Pieter Bruegel the Elder & The Witch of Malleghem & published 1559 & \href{https://artsandculture.google.com/asset/the-witch-of-malleghem-pieter-van-der-heyden-after-pieter-bruegel-the-elder/vQE9PrauEh\_yyw}{Link} \\\midrule
Pieter van der Heyden, Pieter Bruegel the Elder, Hieronymus Cock & Patience (Patientia) & 1557 & \href{https://artsandculture.google.com/asset/patience-patientia-pieter-van-der-heyden-pieter-bruegel-the-elder-hieronymus-cock/oQER1R\_7ykm4xg}{Link} \\\midrule
Francisco de Goya & El Entierro de la Sardina & 1808/1812 & \href{https://artsandculture.google.com/asset/el-entierro-de-la-sardina-francisco-de-goya/AQEEgq6-mXePsA}{Link} \\\midrule
Master of Okolično & Holy Kinship & 1510 & \href{https://artsandculture.google.com/asset/holy-kinship-master-of-okoli\%C4\%8Dno/GwH4shxUzNoo9w}{Link} \\\midrule
Maarten van Heemskerck & The Gods of the Olympus & 1556 & \href{https://artsandculture.google.com/asset/the-gods-of-the-olympus-maarten-van-heemskerck/XwFDbQC5UuACCA}{Link} \\\midrule
Gerrit van Honthorst & Apollo and Diana & 1628 & \href{https://artsandculture.google.com/asset/apollo-and-diana-gerrit-van-honthorst/IAFIUn\_4-HJwQA}{Link} \\\midrule
Pieter Bruegel the Elder & Children’s Games & 1560 & \href{https://artsandculture.google.com/asset/children\%E2\%80\%99s-games-pieter-bruegel-the-elder/CQEeZWQPOI2Yjg}{Link} \\\midrule
Hendrick Avercamp & Skating Scene & 1620s & \href{https://artsandculture.google.com/asset/skating-scene-hendrick-avercamp/ZAHph9BQgNNnhg}{Link} \\\midrule
Pieter Bruegel the Elder & The Adoration of the Magi & undated & \href{https://artsandculture.google.com/asset/the-adoration-of-the-magi-pieter-bruegel-the-elder/\_QFK7xQhJb849A}{Link} \\\midrule
Bol, Ferdinand & Consul Titus Manlius Torquatus Orders the Beheading of his Son & 1661-1663 & \href{https://artsandculture.google.com/asset/consul-titus-manlius-torquatus-orders-the-beheading-of-his-son-bol-ferdinand/7QEgBB09erXRZw}{Link} \\\midrule
Balthasar van der Ast & Still Life with Basket of Fruit & 1622 & \href{https://artsandculture.google.com/asset/still-life-with-basket-of-fruit-balthasar-van-der-ast/4wFn66\_LLgSVWA}{Link} \\\midrule
Pieter Bruegel the Elder, Philips Galle, Hieronymus Cock & The Alchemist & after 1558 & \href{https://artsandculture.google.com/asset/the-alchemist-pieter-bruegel-the-elder-philips-galle-hieronymus-cock/cAHvccLsuc5Vbw}{Link} \\\midrule
Claesz., Pieter & Still Life with a Turkey Pie & 1627 & \href{https://artsandculture.google.com/asset/still-life-with-a-turkey-pie-claesz-pieter/YwEFe\_qO-6us9g}{Link} \\\midrule
Pieter Bruegel the Elder  & The Dirty Bride or the Marriage of Mopsus and Nisa & 1570 & \href{https://artsandculture.google.com/asset/the-dirty-bride-or-the-marriage-of-mopsus-and-nisa-pieter-bruegel-the-elder-netherlandish-breda-ca-1525\%E2\%80\%931569-brussels/pwFfxBwdFYQLoA}{Link} \\\midrule
Western artist painted from the deck of HMS Vulcan & British and French fleets in Victoria Harbour & 1860 & \href{https://artsandculture.google.com/asset/british-and-french-fleets-in-victoria-harbour-western-artist-painted-from-the-deck-of-hms-vulcan/JQFo\_6zy9MJsBA}{Link} \\\midrule
Pieter Bruegel the Elder & Hunters in the Snow (Winter) & 1565 & \href{https://artsandculture.google.com/asset/hunters-in-the-snow-winter-pieter-bruegel-the-elder/WgFmzFNNN74nUg}{Link} \\\midrule
Philips Galle, Pieter Bruegel the Elder, Hieronymus Cock & Justice (Justicia) from The Virtues & ca.~1559–60 & \href{https://artsandculture.google.com/asset/justice-justicia-from-the-virtues-philips-galle-pieter-bruegel-the-elder-hieronymus-cock/zgGGQMw4OinqhQ}{Link} \\\midrule
Giulio Romano & Chamber of the Giants - Ceiling & 1532-1534 & \href{https://artsandculture.google.com/asset/chamber-of-the-giants-ceiling-giulio-romano/1AE7K-84t2gYxg}{Link} \\\midrule
Pieter Bruegel the Elder, Pieter van der Heyden, Hieronymus Cock & The Descent of Christ Into Limbo & ca.~1561 & \href{https://artsandculture.google.com/asset/the-descent-of-christ-into-limbo-pieter-bruegel-the-elder-pieter-van-der-heyden-hieronymus-cock/XgEUYeoADYxG1A}{Link} \\\midrule
Hendrick Avercamp & Winter Landscape & 1600/1620 & \href{https://artsandculture.google.com/asset/winter-landscape-hendrick-avercamp/RgHh4vB0JY3VjA}{Link} \\\midrule
Arcadi Mas i Fondevila & The Corpus Christi procession & 1887 & \href{https://artsandculture.google.com/asset/the-corpus-christi-procession-arcadi-mas-i-fondevila/\_AHEDI-Jao4F2g}{Link} \\\midrule
Unknown & Winter Scene on a Frozen Canal & 1620 & \href{https://artsandculture.google.com/asset/winter-scene-on-a-frozen-canal/mQG53metwNY8fQ}{Link} \\\midrule
Caroline Le Souef & Home Life of the Victorian Aborigines & 1895-1895 & \href{https://artsandculture.google.com/asset/home-life-of-the-victorian-aborigines-caroline-le-souef/8gFHQOS4B3uK1A}{Link} \\\midrule
Andrej Janez Herrlein & Ljubljana Šempeter & 1798 & \href{https://artsandculture.google.com/asset/ljubljana-\%C5\%A0empeter-andrej-janez-herrlein/5wEj\_34z1BhW7Q}{Link} \\\midrule
Edward Roper & Gold Diggings, Ararat & 1855-1860 & \href{https://artsandculture.google.com/asset/gold-diggings-ararat-edward-roper/sQHsGCuj8hmNyQ}{Link} \\\midrule
Pieter Bruegel the Elder & Massacre of the Innocents & 1565-1567 & \href{https://artsandculture.google.com/asset/massacre-of-the-innocents-pieter-bruegel-the-elder/qgGZ6pq1mTaabw}{Link} \\\midrule
Lucas van Leyden & Christ presented to the people & 1510 & \href{https://artsandculture.google.com/asset/christ-presented-to-the-people-lucas-van-leyden/-QFxcMr9nqyKXw}{Link} \\\midrule
Hans Bol & Goose Snatching & 1560/1580 & \href{https://artsandculture.google.com/asset/goose-snatching-hans-bol/jwGMDO77O1PjoA}{Link} \\\midrule
Edouard Hildebrand & Menino com patos & 1800/1800 & \href{https://artsandculture.google.com/asset/menino-com-patos-edouard-hildebrand/KwERhHa6TDO2eQ}{Link} \\\midrule
Pieter Bruegel the Elder & The Census at Bethlehem & 1566 & \href{https://artsandculture.google.com/asset/the-census-at-bethlehem-pieter-bruegel-the-elder/JwGxiyxYTZZEog}{Link} \\\midrule
Frans Huys & Ice Skating before the Gate of Saint George in Antwerp & 1558 & \href{https://artsandculture.google.com/asset/ice-skating-before-the-gate-of-saint-george-in-antwerp-frans-huys/VgGhfGPn2VC-UA}{Link} \\\midrule
Pieter Aertsen & A Meat Stall with the Holy Family Giving Alms & 1551 & \href{https://artsandculture.google.com/asset/a-meat-stall-with-the-holy-family-giving-alms-0000/fgF8j5tB3UFgAg}{Link} \\\midrule
Philip Galle after Pieter Bruegel the Elder & Fortitude & published 1559 & \href{https://artsandculture.google.com/asset/fortitude-philip-galle-after-pieter-bruegel-the-elder/pwFr-l0URpLD4Q}{Link} \\\midrule
Lucas Cranach the Elder & The Crucifixion & 1506-1520 & \href{https://artsandculture.google.com/asset/the-crucifixion-lucas-cranach-the-elder/5AEhErWK5jWo7g}{Link} \\\midrule
Bol, Ferdinand & Aeneas Crowning Cloanthus & ca.~1661-1663 & \href{https://artsandculture.google.com/asset/aeneas-crowning-cloanthus-bol-ferdinand/7QEdHirCaItWKQ}{Link} \\\midrule
Pelegrí Clavé i Roqué & Jacob receives the bloody tunic of his son Joseph & 1842 & \href{https://artsandculture.google.com/asset/jacob-receives-the-bloody-tunic-of-his-son-joseph-pelegr\%C3\%AD-clav\%C3\%A9-i-roqu\%C3\%A9/VgEL15BVsSWu8w}{Link} \\\midrule
Aertsen, Pieter & The Egg Dance & 1552 & \href{https://artsandculture.google.com/asset/the-egg-dance-0001/OAHpfjNpikwVqg}{Link} \\\midrule
Unknown & The Village Wedding & 1653 & \href{https://artsandculture.google.com/asset/the-village-wedding/DwG8I0UTzRQv\_g}{Link} \\\midrule
Joos van Craesbeeck & The Temptation of Saint Anthony & 1650 & \href{https://artsandculture.google.com/asset/the-temptation-of-saint-anthony-joos-van-craesbeeck/5QFZrP\_A46ZgQw}{Link} \\\midrule
Pieter Bruegel the Elder & Winter Landscape with Skaters and Birds Trap & 1565 & \href{https://artsandculture.google.com/asset/winter-landscape-with-skaters-and-birds-trap-pieter-bruegel-the-elder/DAFq10uBnWXIng}{Link} \\\midrule
Caroline Le Souef & Native Fight on the Lower Goulburn River in 1842 & 1895-1895 & \href{https://artsandculture.google.com/asset/native-fight-on-the-lower-goulburn-river-in-1842-caroline-le-souef/JwEvRx6VfwXq7A}{Link} \\\midrule
Joachim Wtewael & The Annunciation to the Shepherds & 1606 & \href{https://artsandculture.google.com/asset/the-annunciation-to-the-shepherds-joachim-wtewael/zwF6\_aMYatvg1A}{Link} \\\midrule
Joachim Beuckelaer  & Woman Selling Vegetables & second half of 16th century & \href{https://artsandculture.google.com/asset/woman-selling-vegetables-joachim-beuckelaer-antwerp-1533-\%E2\%80\%93-1575/3QFkFPRqcu6FVA}{Link} \\\midrule
Unknown & Dragon boat racing at Spring Festival & 1860 & \href{https://artsandculture.google.com/asset/dragon-boat-racing-at-spring-festival/QQF0Dx397nSYew}{Link} \\\midrule
Karl Brullov & The Last Day of Pompeii & 1830/1833 & \href{https://artsandculture.google.com/asset/the-last-day-of-pompeii-karl-brullov/tAFrCGFUhXM8Jg}{Link} \\\midrule
Caroline Le Souef & Corroboree on the Goulburn River & 1895-1895 & \href{https://artsandculture.google.com/asset/corroboree-on-the-goulburn-river-caroline-le-souef/zwGqvxaNovhdDw}{Link} \\\midrule
Pieter Bruegel the Elder, Hieronymus Cock, Philips Galle & Hope (Spes) from The Virtues & ca.~1559–60 & \href{https://artsandculture.google.com/asset/hope-spes-from-the-virtues-pieter-bruegel-the-elder-hieronymus-cock-philips-galle/ngEPU0aWppthwA}{Link} \\\midrule
Steen, Jan Havicksz. & The Feast of St Nicholas & 1665-1668 & \href{https://artsandculture.google.com/asset/the-feast-of-st-nicholas-steen-jan-havicksz/HQH272DWHOl8aA}{Link} \\\midrule
Hans Memling & Virgin and Child with Saints Catherine of Alexandria and Barbara & early 1480s & \href{https://artsandculture.google.com/asset/virgin-and-child-with-saints-catherine-of-alexandria-and-barbara-hans-memling/0AEHESqtvewRHA}{Link} \\\midrule
Gerard Dou & The Quack & 1652 & \href{https://artsandculture.google.com/asset/the-quack-gerard-dou/ugGEqedNalXhvg}{Link} \\\midrule
Baines, Thomas & Forces under General Cathcart crossing the Orange river to attack Moshesh 1852 & 1854 & \href{https://artsandculture.google.com/asset/forces-under-general-cathcart-crossing-the-orange-river-to-attack-moshesh-1852-baines-thomas/tgHyazTHMtT5IQ}{Link} \\\midrule
Namcheong  & Whampoa's earliest mud dock, Canton  & ca.~1850s & \href{https://artsandculture.google.com/asset/whampoa-s-earliest-mud-dock-canton-c-1850s-namcheong-fl-1845-1880s/zQESxCUJvLv54Q}{Link} \\\midrule
Jan Fyt & Still life with parrot & ca.~1645 & \href{https://artsandculture.google.com/asset/still-life-with-parrot-jan-fyt/WAGv5yeMgSurVg}{Link} \\\midrule
Oldmeadow, William H. & Reffley Spring & 1818 & \href{https://artsandculture.google.com/asset/reffley-spring-oldmeadow-william-h/QQHQQdeyYHSlpA}{Link} \\\midrule
Bartholomeus van Bassen, Esaias van den Velde & Renaissance Interior with Banqueters & 1618/1622 & \href{https://artsandculture.google.com/asset/renaissance-interior-with-banqueters-bartholomeus-van-bassen-esaias-van-den-velde/4AFf1S7jP-vDMQ}{Link} \\\midrule
Gift of Mr. Antony J. Hardy & Bamboo Town and Anchorage at Whampoa Island & 1860 & \href{https://artsandculture.google.com/asset/bamboo-town-and-anchorage-at-whampoa-island-gift-of-mr-antony-j-hardy/9QHgy\_4j\_RCB2g}{Link} \\\midrule
Unknown & A handscroll painting of the porcelain production process (left half) & early 19th century & \href{https://artsandculture.google.com/asset/a-handscroll-painting-of-the-porcelain-production-process-early-19th-century-left-half/fAHU6XhS4fSAuQ}{Link} \\\midrule
Pieter Brueghel, Pieter van der Heyden, Hieronymus Cock   & Luxuria, uno de los siete vicios & 1558 & \href{https://artsandculture.google.com/asset/luxuria-uno-de-los-siete-vicios-pieter-brueghel-brueghel-1520-30-bruselas-1569-pieter-van-der-heyden-pa\%C3\%ADses-bajos-ca-1525\%E2\%80\%93amberes-flandes-hoy-b\%C3\%A9lgica-1569-grabador-and-hieronymus-cock-amberes-flandes-hoy-b\%C3\%A9lgica-ca-1510\%E2\%80\%931570-editor/RwG8kgPCys8J-A}{Link} \\\midrule
Pieter Brueghel & La soberbia Serie de los siete pecados capitales & 1558 & \href{https://artsandculture.google.com/asset/la-soberbia-serie-de-los-siete-pecados-capitales-pieter-brueghel/TgHceZyLmd3EYA}{Link} \\\midrule
Pieter Bruegel the Elder  & Landscape with the Fall of Icarus & undated & \href{https://artsandculture.google.com/asset/landscape-with-the-fall-of-icarus-pieter-bruegel-the-elder-after/6gGkgMwPyiEqUQ}{Link} \\\midrule
Pieter Bruegel the Elder, Pieter van der Heyden, Hieronymus Cock & The Last Judgment & 1558 & \href{https://artsandculture.google.com/asset/the-last-judgment-pieter-bruegel-the-elder-pieter-van-der-heyden-hieronymus-cock/hwH93qLb5\_AY3w}{Link} \\\midrule
Hans Memling & The Annunciation & 1480–89 & \href{https://artsandculture.google.com/asset/the-annunciation-hans-memling/mAH0lj4CEmyW\_w}{Link} \\\midrule
James Taylor & Panorama du Port Jackson et de la ville de Sidney & 1820/1825 & \href{https://artsandculture.google.com/asset/panorama-du-port-jackson-et-de-la-ville-de-sidney-ca-1821-drawn-by-tailor-i-e-taylor-with-key-to-buildings-james-taylor/mQEXVCU28f42vw}{Link} \\\midrule
Avercamp, Hendrick & Frolicking on a Frozen Canal in a Town & ca.~1615-1620 & \href{https://artsandculture.google.com/asset/frolicking-on-a-frozen-canal-in-a-town-avercamp-hendrick/FgGJNJ33DQ-zaA}{Link} \\\midrule
After Pieter Bruegel the Elder  & Gluttony (Gula) from The Seven Deadly Sins & 1558 & \href{https://artsandculture.google.com/asset/gluttony-gula-from-the-seven-deadly-sins-after-pieter-bruegel-the-elder-netherlandish-breda-ca-1525\%E2\%80\%931569-brussels/LAEs2sZfnV8j\_Q}{Link} \\\midrule
Felip Massó i Falp & The Procession of St. Bartolomew & 1884 & \href{https://artsandculture.google.com/asset/the-procession-of-st-bartolomew-felip-mass\%C3\%B3-i-falp/qQHjCbMqzCI17A}{Link} \\\midrule
Pieter Bruegel the Elder & The Dutch Proverbs & 1559 & \href{https://artsandculture.google.com/asset/proverbs-pieter-bruegel-the-elder/CwGkTFwDGMmq9w}{Link} \\\midrule
Maertan Van Heemskerck & Concert of Apollo and the Muses on Mount Helicon & 1565 & \href{https://artsandculture.google.com/asset/concert-of-apollo-and-the-muses-on-mount-helicon-maertan-van-heemskerck/1wFwfN0AzbJ3xA}{Link} \\\midrule
David Teniers the Younger & The Surgeon & 1670s & \href{https://artsandculture.google.com/asset/the-surgeon-david-teniers-the-younger/ogG\_vPMwkyxUKQ}{Link} \\
\bottomrule
\end{longtable}

\end{document}